\documentclass[review]{elsarticle}

\usepackage{amsmath}
\usepackage{array}
\usepackage{afterpage}
\usepackage{titlesec}
\usepackage{enumitem}
\usepackage{multirow}
\usepackage{algorithm}
\usepackage{algorithmicx}
\usepackage[noend]{algpseudocode}
\usepackage{hyperref}
\makeatletter
\def\BState{\State\hskip-\ALG@thistlm}
\makeatother

\journal{Neural Networks}

\begin{document}

\begin{frontmatter}

\title{TRk-CNN: Transferable Ranking-CNN for image classification of glaucoma, glaucoma suspect, and normal eyes}

\author[KAIST]{Tae Joon Jun}
\author[KUM]{Youngsub Eom}
\author[KAIST]{Dohyeun Kim}
\author[KUM2]{Cherry Kim}
\author[KUM]{Ji-Hye Park}
\author[KAIST]{Hoang Minh Nguyen}
\author[KAIST]{Daeyoung Kim\corref{CORRESPONDING}}
\cortext[CORRESPONDING]{Corresponding author}
\ead{kimd@kaist.ac.kr}

\address[KAIST]{
	School of Computing,
    Korea Advanced Institute of Science and Technology,\\
    34141 Daejeon, Republic of Korea}
    
\address[KUM]{
	Department of Ophthalmology,
    Korea University College of Medicine\\
    02841, Seoul, Republic of Korea.}
    
\address[KUM2]{
	Department of Radiology,
    Korea University College of Medicine\\
    02841, Seoul, Republic of Korea.}

\begin{abstract}
% 수정 필요
In this paper, we proposed Transferable Ranking Convolutional Neural Network (TRk-CNN) that can be effectively applied when the classes of images to be classified show a high correlation with each other. The multi-class classification method based on the softmax function, which is generally used, is not effective in this case because the inter-class relationship is ignored. Although there is a Ranking-CNN that takes into account the ordinal classes, it cannot reflect the inter-class relationship to the final prediction. TRk-CNN, on the other hand, combines the weights of the primitive classification model to reflect the inter-class information to the final classification phase. We evaluated TRk-CNN in glaucoma image dataset that was labeled into three classes: normal, glaucoma suspect, and glaucoma eyes. Based on the literature we surveyed, this study is the first to classify three status of glaucoma fundus image dataset into three different classes. We compared the evaluation results of TRk-CNN with Ranking-CNN (Rk-CNN) and multi-class CNN (MC-CNN) using the DenseNet as the backbone CNN model. As a result, TRk-CNN achieved an average accuracy of 92.96\%, specificity of 93.33\%, sensitivity for glaucoma suspect of 95.12\% and sensitivity for glaucoma of 93.98\%. Based on average accuracy, TRk-CNN is 8.04\% and 9.54\% higher than Rk-CNN and MC-CNN and surprisingly 26.83\% higher for sensitivity for suspicious than multi-class CNN. Our TRk-CNN is expected to be effectively applied to the medical image classification problem where the disease state is continuous and increases in the positive class direction.
\end{abstract}


\end{frontmatter}

%\linenumbers

\section{Introduction}
\label{ch2:introduction}
The rapid development of deep learning technologies, especially convolutional neural network (CNN), is now considered to be a cutting-edge methodology for classifying medical images. The vast majority of recent medical image analysis literature uses CNN-based methodologies. The main reason CNN is effective in medical image analysis is that CNN is trained end-to-end. In other words, CNN's automated feature extraction process is more effective than traditional handcrafted feature extraction methods. However medical image classes are distinguished from general image classes. That is, the classes of medical images have a strong correlation with each other. In particular, there are innumerable intermediate states between the negative class, which is classified as normal, and the positive class, which is classified as disease. In addition, the negative class proceeds to a positive class in the direction of increasing the inherent characteristic. This characteristic depends on the type of medical image to be classified. For example, the cancer staging using the TNM system includes the size of the tumor \cite{ch2_tnm}, and in a cataract patient, the degree of turbidity of the ocular lens may be increased \cite{ch2_locs}.

Thus, the actual disease state is continuous and increases in the positive class direction. However, when the medical image is taken and the physician makes a decision, the class of the medical image is determined based on a certain point on the continuous line. Therefore, depending on the disease, the intermediate class of the medical image may be defined by the physician, not dichotomous separation into normal and disease. Glaucoma is a representative example of such diseases. The reason is that glaucoma should be treated appropriately before advanced stages where they are already positive, and the disease is worsening over a long period, it is necessary to observe persistent intermediate conditions. Glaucoma is an eye disease that causes narrowed vision and eventually leads to blindness, which is caused by various reasons such as elevated intra-ocular pressure (IOP) or blood circulation disorder \cite{ch2_glaucomadef}. Once glaucoma is diagnosed, it needs constant management for a lifetime, and the damaged vision is not restored. Therefore, early detection and treatment of glaucoma is the best prevention, but the optic nerve damage caused by glaucoma gradually develops, and when symptoms appear, the disease progresses considerably. In addition, since it is not easy to confirm glaucoma early, various tests including IOP measurement, optic nerve head examination, and anterior chamber angle examination are conducted and the results are combined to determine the existence of glaucoma.

As a result, recent glaucoma fundus image dataset includes the glaucoma suspect class and there are several existing studies that detect glaucoma using machine learning methods. Most of them use multi-class classification method that uses CNN as a classifier and utilizes the output values of softmax function. The literature on classification of glaucoma from fundus images will be discussed in more detail in the related work section. Although such machine learning based eye disease classification studies show reasonable performance, this multi-class classification method ignores inter-class information of eye diseases. In addition, in the binary classification problem of classifying normal and glaucoma, the addition of suspect class results in poor overall classification performance. In other words, in the case of diseases that show a sequential relationship among medical image classes, a method that can classify them considering the inter-class relationship is required.

Therefore, we propose a Transferable Ranking-CNN (TRk-CNN) for glaucoma detection considering information between three different fundus image classes. TRk-CNN consists of the following steps: primitive classification, region of interest (ROI) extraction, and final classification. Primitive classification follows the general Ranking-CNN \cite{ch2_rankcnn} procedure. Ranking-CNN will be described in detail in the later sections. Briefly, it is a method of aggregating results of \textit{N} - 1 binary classifiers to classify \textit{N} number of ordinal classes. More specifically, when classifying \textit{N} ordinal classes, \textit{k}-th sub-classifier determine whether the predicted class is higher than the class \textit{k} which ranges between 1 to \textit{N} - 1. The difference from the original Ranking-CNN in primitive classification is that there are no fully-connected layers at the top-layers of the CNN classifier that performs binary classification. As a result of the primitive classification, we get the Class Activation Map (CAM) \cite{ch2_cam} for the predicted class. The CAM will also be discussed in detail later, but in a nutshell, it includes the importance of which spatial location in the input image highly affects the final prediction. The CAMs obtained from the \textit{N}-1 sub-classifiers are combined into a single ROI based on the inter-class distance metrics definition, and the process of extracting the ROI and combining it with the original input is processed in the ROI extraction step. The new input, combined with the ROI, is used as an input to the final classification step. In this step, the final class is predicted through a sophisticated classification process including a fully-connected layer. 

We evaluated TRk-CNN in glaucoma image dataset that was collected and labeled from Korea University Medical Center. Glaucoma dataset was labeled into three classes: normal, glaucoma suspect, and glaucoma eyes. Based on the literature we surveyed, this study is the first to classify three status of glaucoma fundus image dataset into three different classes. We compared the evaluation results of TRk-CNN with multi-class CNN (MC-CNN) and Ranking-CNN (Rk-CNN) using the DenseNet \cite{ch23_densenet} as the backbone CNN model. As a result, TRk-CNN achieved an average accuracy of 92.96\%, specificity of 93.33\%, sensitivity for glaucoma suspect of 95.12\% and sensitivity for glaucoma of 93.98\%. Based on average accuracy, TRk-CNN is 8.04\% and 9.54\% higher than Rk-CNN and MC-CNN and surprisingly 26.83\% higher for sensitivity for suspicious than MC-CNN.

The major contribution of this work is summarized as follows:
\begin{itemize}
\item  Our proposed TRk-CNN is a method that can be effectively applied when the classes of images to be classified show a high correlation with each other. The multi-class classification method based on the softmax function, which is generally used, is not effective in this case because the inter-class relationship is ignored. Although there is the Ranking-CNN that takes into account the ordinal classes, it cannot reflect the inter-class relationship to the final prediction. TRk-CNN, on the other hand, combines the weights of the primitive classification model to reflect the inter-class information to the final classification phase. Through extensive experiments, we show that TRk-CNN is superior to both multi-class classification method and Ranking-CNN method.
\item We evaluated TRk-CNN in glaucoma fundus images. Glaucoma can be labeled with suspicious states because it is important to find and take proper treatment before the condition becomes severe. We think that this is not a problem specific to glaucoma. Many diseases requiring medical imaging have intermediate states from negative class to positive class. Our TRk-CNN is expected to be effectively applied to those medical image classification problem using CNN.
\end{itemize}

The abstract version of this paper has been published in \cite{ch2_2srank}. Compared with \cite{ch2_2srank}, this paper presents TRk-CNN as a general classification model that can be applied not only to three classes but also to \textit{N} number of classes. We have also noticed that \cite{ch2_2srank} showed an unusually high classification accuracy because the train-set and test-set of primitive and final classification steps are divided based on different random seeds. We have corrected the above error in this paper. In addition, a more robust evaluation was conducted to compare with the results of previous glaucoma detection studies. The rest of this paper is structured as followed. In Section 2, we review the literature using a machine-learning approach that includes deep-learning for glaucoma detection and also briefly review the multi-class classification and Ranking-CNN that is the background of this study. Section 3 explains in detail the three steps of TRk-CNN in the general example of classifying \textit{N} different classes. Section 4 describes the optimal TRk-CNN for glaucoma detection. In Section 5, we evaluate TRk-CNN in glaucoma dataset and compares the result with multi-class CNN and Ranking-CNN results. Finally, we conclude this study in Section 6 and discusses future plans.

%-------------------------------------------------------------------------
\section{Related Work}
\label{ch2:related-work}
\subsection{Glaucoma detection}
\begin{figure}[!tbh]
	\centering
	\includegraphics[width=\textwidth]{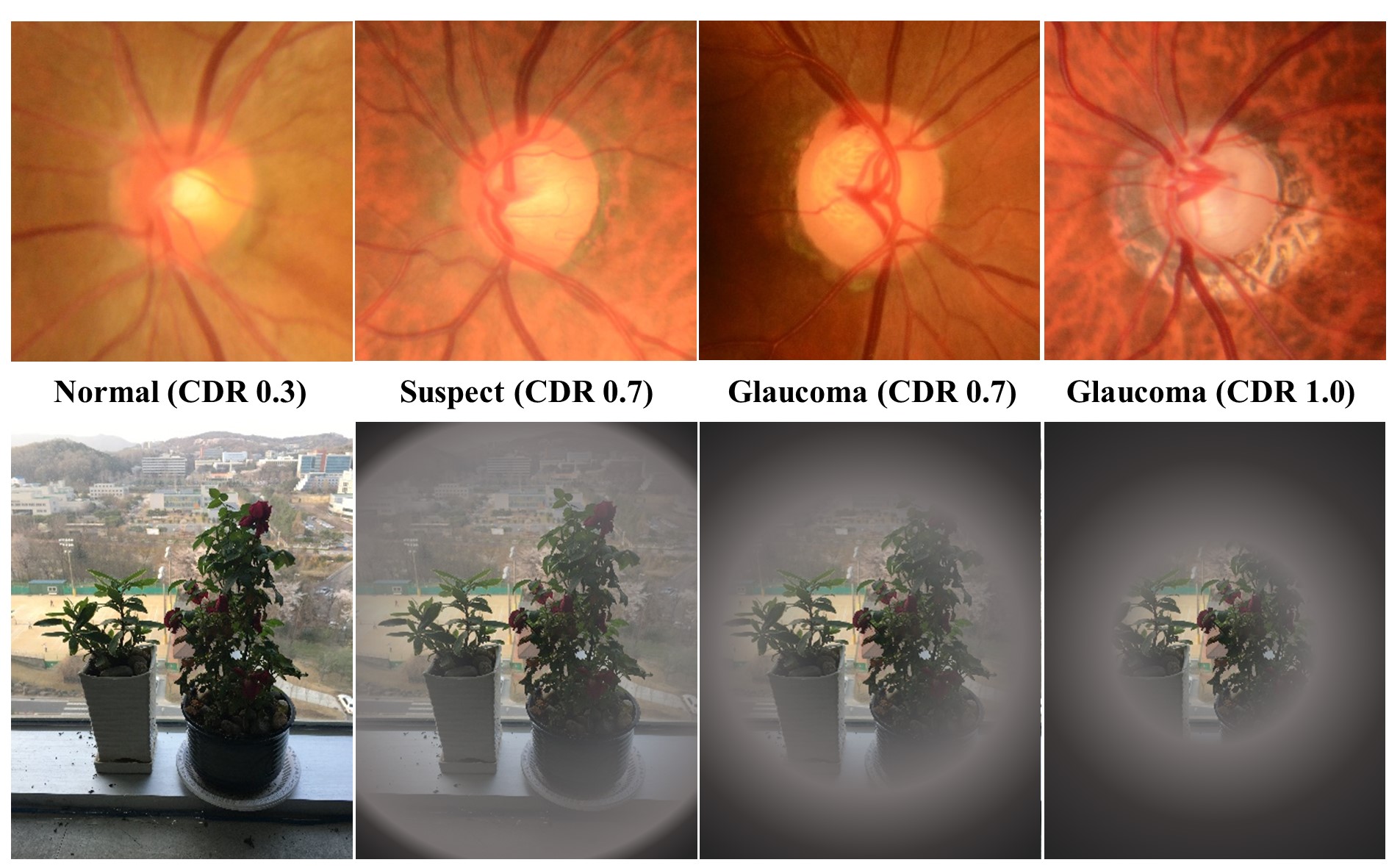}
	\caption{Optic disc changes and visual field loss with normal, glaucoma suspect, and glaucoma eyes}
	\label{img:glaucoma_vision}
\end{figure}

Glaucoma is a disease in which the optic nerve and nerve fiber layers, which play an important role in delivering visual information received from the eye to the brain, are damaged and the visual field becomes narrower. Globally, glaucoma is a major cause of blindness, along with cataracts and diabetic retinopathy, and is one of the most common ophthalmic diseases, with a frequency of 2\% of the total population \cite{ch2_glaucomaworld1} \cite{ch2_glaucomaworld2} \cite{ch2_glaucomaworld3}. In the past, glaucoma generally included increased intra-ocular pressure, but recently, normal tension glaucoma is a very common disease, and the definition of glaucoma has also changed. Primary open-angle glaucoma and normal-tension glaucoma, which account for the vast majority of glaucoma, chronically and slowly damage the optic nerve \cite{ch2_openangle}. As a result, visual field damage progresses, damage to the peripheral vision first occurs, and central vision is often preserved until the end of the period. Therefore in the beginning, there is almost no subjective symptom and symptoms do not appear until glaucoma has progressed to advanced stages. As a result, most of the patients diagnosed with glaucoma are found incidentally through ophthalmologic examination or physical examination regardless of the glaucoma related symptoms. Figure \ref{img:glaucoma_vision} shows the progression of optic disc changes and visual field defects with normal, glaucoma suspect, and glaucoma eyes.
%% 녹내장으로 시야가 결손되는 그림 보여주기

To overcome the difficulties in early diagnosis, applying machine learning methods to classify normal and glaucoma in fundus image have been proposed to play a supporting role in physician's glaucoma diagnosis criteria. 

In 2009, Nayak proposed a method to classify normal and glaucoma with single hidden-layer neural network (ANN) by extracting features such as cup-to-disc ratio (CDR), optic nerve head shift, and ISNT ratio from the fundus image \cite{ch2_nayak}. ISNT ratio is the total area of the blood vessels in the inferior and superior side of the optic disc to the total area of the blood vessels in the nasal and temporal area. Of the 24 normal and 37 glaucoma images, 5 normal and 10 glaucoma images were split into test-set. As a result, the specificity (Sp) was 80\% and the sensitivity (Se) was 100\%. Nayak's work is meaningful in that it extracts features and train them by the neural network, although the number of images is too small.

Bock proposed a method for extracting a probabilistic feature for glaucoma diagnosis from a fundus image called glaucoma risk index (GRI) in 2010 \cite{ch2_bock}. First, they perform pre-processing procedures such as illumination correction, vessel removal, and optic nerve head normalization. Then, Fourier analysis and spline interpolation are applied, and principal component analysis (PCA) is performed to extract features. Finally, the features extracted by the PCA are passed into two-stage support vector machine (SVM) classifier and finally the classifier outputs a GRI indicating the probability for glaucoma. For 575 fundus images consisting of 336 normal and 239 glaucoma, the GRI method showed overall 80\% accuracy with the area under the receiver operating characteristic (ROC) curve (AUC) of 0.88, sensitivity of 73\%, and specificity of 85\%. Bock work is also a representative approach to extract handcrafted features from images and use them as inputs for classifiers such as SVM and neural networks.

In 2011, Acharya proposed a method for extracting higher order spectra (HOS) parameter and texture descriptors from a fundus image and use them as inputs for four different classifiers \cite{ch2_acharya}. Classifiers are SVM, sequential minimal optimization (SMO), naive Bayesian, and random-forest. As a result, random forest classifier showed the best performance with accuracy (Acc) of 91.7\% in 60 fundus images composed of 30 normal and 30 glaucoma eyes. For the same dataset as Acharya's work, Dua proposed a method for extracting energy signatures as a feature by applying a 2-dimensional discrete wavelet transform to fundus images in 2012 \cite{ch2_dua}. Again for the four classifiers including SVM, SMO, naive Bayesian (NB), and random-forest (RF), Dua's work achieved the highest accuracy of 93.33\% in both SVM and SMO classifiers.

From 2015, glaucoma detection studies based on convolutional neural networks have become mainstream with the rapid development of deep learning technology. Chen performed a classification of normal and glaucoma fundus images using CNN in 2015 \cite{ch2_chen}. Chen designed the AlexNet \cite{ch23_alexnet} based CNN model, and evaluated with the ORIGA \cite{ch2_origa} and SCES \cite{ch2_sces} fundus image dataset. The ORIGA dataset composed of 168 glaucoma and 482 normal fundus images and SCES dataset contains 1676 fundus images including 46 glaucoma cases. As a result, Chen obtained 0.831 and 0.887 AUC on ORIGA and SCES dataset. Chen's work is meaningful in that it is the first study which applied CNN's end-to-end training to glaucoma detection, deviating from the conventional manual feature extraction method. However, it did not perform better than the existing method because it simply applied CNN and did not refine the sophisticated optimization process. In 2016, Li proposed a method to apply CNN models to the disc region and the original fundus image, respectively, and ensemble the predictions \cite{ch2_li1}. Li used four well known CNN models including AlexNet \cite{ch23_alexnet}, GoogLeNet \cite{ch23_googlenet}, 16-layer VGGNet \cite{ch23_vggnet}, and 19-layer VGGNet \cite{ch23_vggnet}. Evaluated with ORIGA dataset, Li achieved AUC of 0.838. CNN is adopted and considering that the classification was binary classification, performance is not good, and similar to Chen's work, there is a limitation that CNN model optimization is not sophisticated.

In 2018, Fu proposed a disc-aware ensemble network for glaucoma classification \cite{ch2_fu}. U-Net \cite{ch23_unet} was used for disc region segmentation and re-applied the resulting region to the original image to reduce the size of the input. Finally, 50-layer ResNet \cite{ch23_resnet} was applied to fundus images of the various regions including disc region and original fundus images. The evaluation was performed in SCES and Singapore Indian Eye Study (SINDI) \cite{ch2_fu} dataset and showed 0.918 AUC and 0.817 AUC, respectively. SINDI dataset contains a total of 5783 fundus images including 113 glaucoma and 5670 normal eyes. Fu's work has ensured the results by applying CNN to various regions similar to Li's work \cite{ch2_li1}, and the CNN model is well optimized. Also in 2018, Li classified the glaucoma eyes by applying the GoogLeNet to 48116 fundus images, which is the largest number of a dataset in the literature \cite{ch2_li2}. They also labeled the dataset as normal, glaucoma suspect, and glaucoma eyes, same as in our study. Dataset consists of 31745 train-set and 8000 test-set images. The train-set consists of 23433 normal, 2190 glaucoma suspect, and 6122 glaucoma eyes. The test-set consists of 6033 normal, 430 glaucoma suspect, and 1537 glaucoma eyes. However, the evaluation was performed as a binary classification to classify normal and abnormal (glaucoma suspect and glaucoma cases). As a result, they obtained 0.986 AUC, sensitivity of 95.6\%, and specificity of 92\%. 

Overall, none of the studies described above take into account to classify the three continuous classes of normal, glaucoma suspect, and glaucoma eyes. Li's work \cite{ch2_li2} is the only one that labels the fundus image in three classes but performs the binary classification by treating glaucoma suspect and glaucoma as a single positive class. As we will see later in the evaluation, binary classification of fundus images with CNN models of the same structure is 10\% higher overall accuracy than three class classification. Therefore, in order to improve the performance of the normal, glaucoma suspect, and glaucoma classification, TRk-CNN which considers inter-class information is necessary. In addition, TRk-CNN can be effectively applied to the classification of other medical images having intermediate stages between negative and positive cases.

%-------------------------------------------------------------------------

\subsection{Multi-class classification and Ranking-CNN}
The multi-class classification is a method in which the size of the final prediction vector is \textit{N} for \textit{N} number of classes. In addition, the \textit{N} different classes are converted to one-hot-encoding, where the index to which they belong is 1 and the remainder is 0. Generally, in deep learning, the softmax function is applied to the output vector to express as the probability between 0 and 1, although it is not the actual probability, and predict the class with the largest probability as the final class. In this case, the cross entropy of the probability of a class that is a true class becomes a loss, which is an error. Therefore, in the next epoch of training, gradient descent is processed in the direction of reducing this loss. However, when classes are highly related to each other, their inter-class relationship disappears because classes are one-hot-encoded in multi-class classification. Especially, the age prediction problem is where this problem is obvious. For example, in the case of classifying tree, truck, and cat images, there is no problem in classifying \big[1,0,0\big], \big[0,1,0\big], and \big[0,0,1\big] through one-hot-encoding. However, when one-hot-encoding is used to classify 10-year-old, 11-year-old, and 12-year-old face images, the ordinal relationship of the age disappears.

Ranking-CNN was proposed by Chen in for age estimation from human face images \cite{ch2_rankcnn}. Prior to Ranking-CNN, ranking algorithms for machine learning-based age estimation such as Ranking SVM \cite{ch2_ranksvm}, Rank-Boost \cite{ch2_rankboost1} \cite{ch2_rankboost2}, and RankNet \cite{ch2_ranknet} were introduced. Ranking-CNN proposed a ranking algorithm suitable for CNN-based facial age estimation problem. In the case of classifying \textit{N} different ages from images, Ranking-CNN creates \textit{N}-1 sub-CNN models, and each model performs binary classification with one age as a reference point. For example, when predicting the ages of 10 to 19-year-old faces, the first CNN model classifies whether the face age is older than 10 years or not. Similarly, the \textit{i}-th sub-CNN model classifies facial images that are order than \textit{i} years old and continues until the 9-th sub-CNN model. For a single facial image, nine different \big[0,1\big] are output as the result, and the final age is determined based on the sum of these values. The major contribution of Ranking-CNN is that by taking the ordinal relation between ages into consideration, Ranking-CNN is more likely to get smaller estimation errors when compared with multi-class classification approaches \cite{ch2_rankcnn}.

However, since Ranking-CNN considers only the final binary value of the trained sub-CNN models, features extracted during the training of each sub-CNN model cannot be transferred. In addition, age is an ordinal relationship, but the classes of medical data like normal, glaucoma suspect, and glaucoma are not always directly proportional to the class relationship. Therefore, our proposed TRk-CNN can achieve higher accuracy by allowing each sub-CNN model to transfer the extracted high-dimensional features. 

%-------------------------------------------------------------------------
\section{Transferable Ranking-CNN}
\label{ch2:trk-cnn}
\begin{figure}[!tbh]
	\centering
	\includegraphics[width=0.95\textwidth]{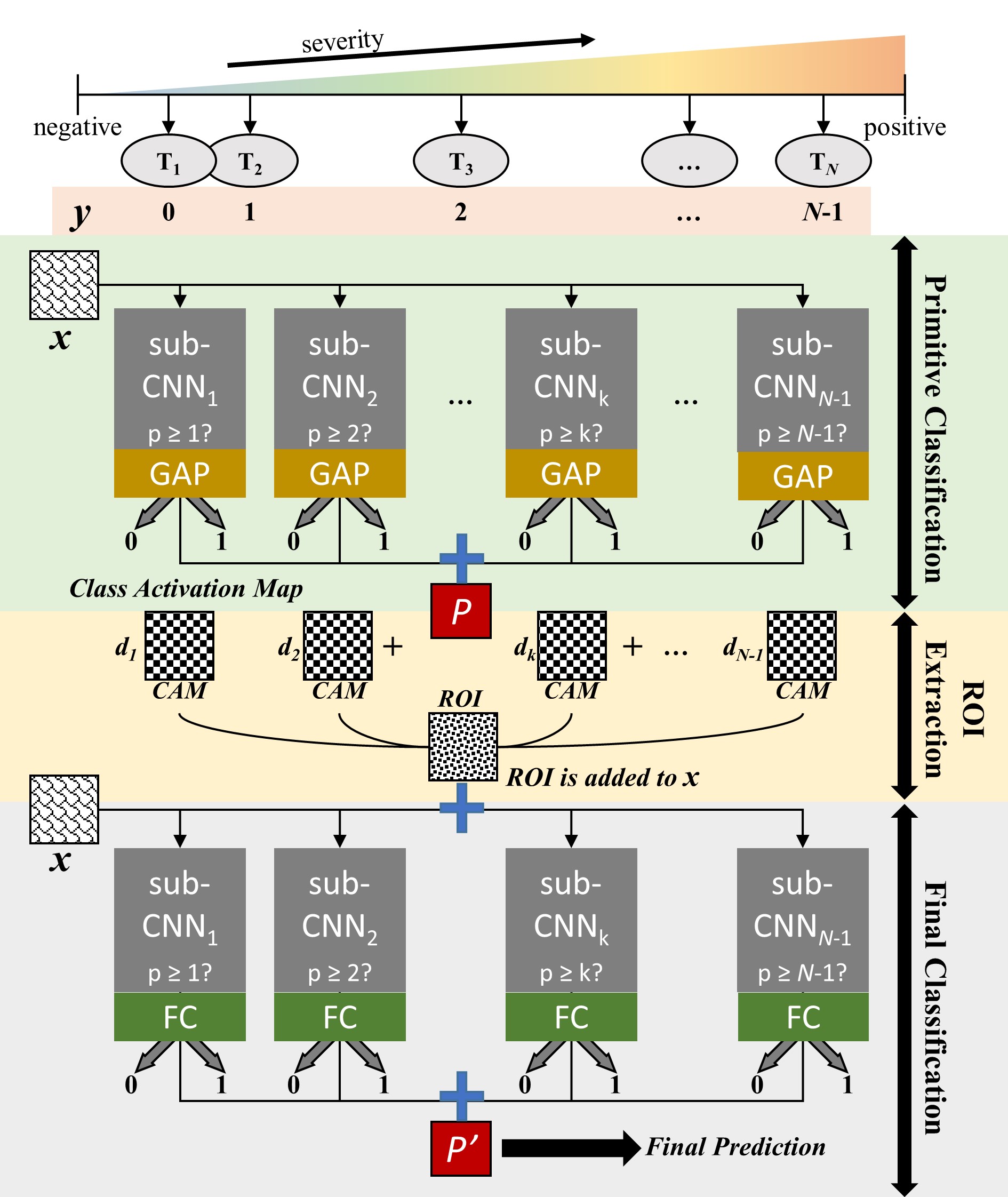}
	\caption{The overall architecture of TRk-CNN}
	\label{img:trk_cnn_all}
\end{figure}

TRk-CNN consists of the following steps: Primitive classification, ROI extraction, and Final classification. Figure \ref{img:trk_cnn_all} shows the overall structure of TRk-CNN including primitive classification, ROI extraction, and final classification steps. 
%전체 TRk-CNN architecture를 하나의 그림으로 나타내자

Primitive classification step follows the general Ranking-CNN procedure and its purpose is to extract the major features of the reference class of each sub-CNN model. The major feature here is that each sub-CNN model should extract different features according to the result of performing a binary classification on a given input image. Therefore, we can not generally use the weight of the last convolutional layer of well known CNN such as VGGNet, GoogLeNet, and ResNet. The reason is that the weight of the last convolutional layer contains the general characteristics of the entire dataset, but of each of the input image, the weight does not include the characteristics of the classification results the sub-CNN model. Features extracted from each sub-CNN model in TRk-CNN should individually represent the characteristics of \textit{N} different classes. In order to satisfy these requirements, Class Activation Map (CAM) is extracted for each input image as a transferable feature of each sub-CNN model. A more detailed description of CAM will be discussed in the later section. In other words, the purpose of primitive classification is to obtain the CAM for input image from each sub-CNN model through training. 

In ROI extraction step, CAM extracted from each sub-CNN model is merged into a single ROI. However, when CAMs are combined through simple summation, the low relevant classes and the high relevant classes are treated equally. So we take into account the association between classes by including the distance function when merging the CAMs. In this paper, we define the distance function assuming that classes have a linear relationship. However, the distance function depends on how the domain expert defines the relationship between classes. For example, a linear relationship is reasonable for age prediction, but it is highly likely that it will not be linear in medical data. As a result, the ROI extraction step is to combine these distance functions with the CAM to create the final ROI of each input image and pass the generated ROI to the final classification step.

Final classification step combines the ROI, which received from the previous ROI extraction stage, with the original image to create a new input for classification. Although there are many possible ways to combine the ROI with the original image, we concatenate the ROIs on the additional channels of the input to preserve the information of the original image. In other words, if the original image has three channels, the number of channels for the new input is now four. We will explain other possible methods in more detail in the later section. Since this step leads to the final prediction, hyper-parameter tuning is strict and regularization is applied more strongly than primitive classification. In addition, the final classification also follows the Ranking-CNN structure and starts to converge from the earlier epoch by loading the pre-trained weights of the model from the primitive classification stage. A detailed description of each stage is provided in the following section.

%-------------------------------------------------------------------------
\subsection{Primitive Classification}
Since primitive classification is almost similar to general Ranking-CNN, we will explain it with the notation from the original paper almost as it is. Let us first assume that the number of classes in the image dataset \textit{X} we want to classify is \textit{N}. Each class is labeled from 0 to \textit{N}-1 depending on the direction in which the state of the class is increasing. Here, the examples of increasing refers to the age in the facial age estimation problem and severity of lesion in the medical image classification problem. When the arbitrary sample belonging to dataset \textit{X} is \textit{x}, the corresponding label of \textit{x} is \textit{y}, where \textit{y} $\in$ \{0, 1, ..., \textit{N}-1\}. As described in the related work section, Ranking-CNN creates \textit{N}-1 number of sub-CNN models to classify dataset \textit{X}. The role of \textit{k}-th sub-CNN model is to perform binary classification in dataset \textit{X} based on reference class \textit{k}. If \textit{x} is classified to be greater than or equal to \textit{k}, the output is 1 and if it is classified to be smaller than \textit{k}, the output is 0. After training \textit{k}-th sub-CNN model, dataset \textit{X} is divided into two subsets as shown below.
\begin{equation}\label{(equ2.1)}
   \begin{split}
	\textit{X}\textsuperscript{0}\textsubscript{k} & = 
	\{(\textit{x}, 0) | \textit{y} < k\}\\
	\textit{X}\textsuperscript{1}\textsubscript{k} & = 
	\{(\textit{x}, 1) | \textit{y} \geq k\}
	\end{split}
\end{equation}
Let the output value of the \textit{k}-th sub-CNN model for arbitrary input \textit{x} is \textit{p}\textsubscript{k}(\textit{x}) where the value is 0 or 1. The role of primitive classification here is to optimize each \textit{k}-th sub-CNN model to minimize the binary classification error. After error is reduced enough, we aggregate the \textit{p}\textsubscript{k}(\textit{x}) of all sub-CNN models for arbitrary input \textit{x} as follows.
\begin{equation}\label{(equ2.2)}
	\textit{P}(\textit{x}) = 
	\sum_{\textit{k}=1}^{\textit{N}-1}\textit{p}\textsubscript{k}(\textit{x})
\end{equation}
where \textit{P}(\textit{x}) corresponds to the predicted class of primitive classification for  arbitrary input \textit{x}. The important point here is that the class we deliver to the ROI extraction step should be the predicted class \textit{P}(\textit{x}), not the actual class \textit{y}. The reason is that if the ROI is created through an actual class, we can not generate the ROI for test-set where the actual class is only available in the final evaluation phase. In other words, if ROI is created with an actual class, test-set cannot be evaluated after the final classification step because the input is an original image without ROI. In the primitive classification, the fully-connected layer cannot come after the last convolutional layer, and the class classifier should follow immediately after the Global Average Pooling (GAP) layer. The reason for this will be explained in detail in the next section, ROI extraction step. Algorithm \ref{algo:pri_cla} provides the entire process of training and validation procedure of primitive classification step.
%% Primitive Classification Algorithm

\begin{algorithm}
\caption{Primitive Classification}\label{algo:pri_cla}
\begin{algorithmic}[1]
\Procedure{Training Procedure}{}
\For{$ k=1\hspace{2mm}to\hspace{2mm}N-1 $}
\State $ \textbf{initialize}\hspace{2mm}\textit{k-th sub-CNN} $
\EndFor
\BState \emph{top}:
\For{$ k=1\hspace{2mm}to\hspace{2mm}N-1 $}
\State $\textit{X}\textsuperscript{0}\textsubscript{k} = \{(\textit{x}, 0) | \textit{y} < k\}$
\State $\textit{X}\textsuperscript{1}\textsubscript{k} = \{(\textit{x}, 1) | \textit{y} \geq k\}$
\State $ \textbf{fine-tune}\hspace{2mm}\textit{k-th sub-CNN} $
\EndFor
\If {$ \textit{not converged} $}
\State $ \textbf{goto}\hspace{2mm}top $
\EndIf
\EndProcedure
\Procedure{Prediction Procedure}{}
\For{$ k=1\hspace{2mm}to\hspace{2mm}N-1 $}
\State $ \textit{p\textsuperscript{k}(x)} \gets \textit{k-th sub-CNN} $
\EndFor
\State $ \textit{P(x)} \gets \sum_{\textit{k}=1}^{\textit{N}-1}\textit{p\textsuperscript{k}(x)} $
\EndProcedure
\end{algorithmic}
\end{algorithm}
%-------------------------------------------------------------------------
\subsection{ROI Extraction}
The outputs from primitive classification step to ROI extraction step are the predicted value \textit{P}(\textit{x}) for input \textit{x} and the weights of trained sub-CNN models. In the previous section, we explained that the Global Average Pooling layer comes after the last convolutional layer of each sub-CNN model, and the fully-connected layers can not. The reason is that Class Activation Map is the feature of input \textit{x} that we want to extract from each sub-CNN model and it requires GAP layer directly after the last convolutional layer. CAM is a concept introduced by Zhou, and it schematically shows which spatial location of the input image played an important role when classified as the final class \cite{ch2_cam}. In general, combining CAM with original input in case of training a single CNN model is not expected to have a great effect on performance, but when combining results trained by multiple CNN models, such as TRk-CNN, CAM can be used to transfer important features between CNN models.

Lets assume that \textit{f}\textsubscript{m}\textsuperscript{k}(\textit{i},\textit{j}) is the activation result of filter \textit{m} $\in$ \{1, 2, ..., \textit{n}\} in the last convolutional layer of \textit{k}-th sub-CNN model at spatial location (\textit{i},\textit{j}) of filter \textit{m}. The size of filter \textit{m} depends on the pooling policy of the sub-CNN model. Suppose the sub-CNN model performs stride 2 pooling, which is a general situation, for \textit{l} number of times. When the size of input \textit{x} is \textit{h} x \textit{h}, the size of filter \textit{m} becomes \textit{h}/2\textsuperscript{\textit{l}} x \textit{h}/2\textsuperscript{\textit{l}}. Finally, the result \textit{F}\textsubscript{m}\textsuperscript{k} obtained from applying GAP layer to filter \textit{m} can be expressed by the following equation.
\begin{equation}\label{(equ2.3)}
	\textit{F}\textsubscript{m}\textsuperscript{k} = 
	\sum_{\textit{(i,j=1)}}^{\textit{h}/2\textsuperscript{\textit{l}}}\textit{f}\textsubscript{m}\textsuperscript{k}(\textit{i},\textit{j})
\end{equation}

From the primitive classification step, predicted class \textit{p}(\textit{x}) is either 0 or 1. Thus, if the predicted class \textit{p}(\textit{x}) is 1 in \textit{k}-th sub-CNN model, the input \textit{S}\textsubscript{1}\textsuperscript{k} for the softmax layer as final prediction can be expressed by the following equation.
\begin{equation}\label{(equ2.4)}
	\textit{S}\textsubscript{1}\textsuperscript{k} = 
	\sum_{\textit{m=1}}^{n}\textit{w}\textsubscript{m}\textsuperscript{1}\textit{F}\textsubscript{m}\textsuperscript{k}
\end{equation}
where \textit{w}\textsubscript{m}\textsuperscript{1} represents the weights between \textit{m}-th node of GAP layer and class 1 node in softamx layer and \textit{n} refers the total number of filters in the last convolutional layer. Substituting \textit{F}\textsubscript{m}\textsuperscript{k} with equation \ref{(equ2.3)} into \textit{S}\textsubscript{1}\textsuperscript{k} yields the following equation.
\begin{equation}\label{(equ2.5)}
  \begin{split}
	\textit{S}\textsubscript{1}\textsuperscript{k} & = \sum_{\textit{m=1}}^{n}\textit{w}\textsubscript{m}\textsuperscript{1}\textit{F}\textsubscript{m}\textsuperscript{k}\\
       & = \sum_{\textit{m=1}}^{n}\textit{w}\textsubscript{m}\textsuperscript{1}\sum_{\textit{(i,j=1)}}^{\textit{h}/2\textsuperscript{\textit{l}}}\textit{f}\textsubscript{m}\textsuperscript{k}(\textit{i},\textit{j})\\
       & = \sum_{\textit{(i,j=1)}}^{\textit{h}/2\textsuperscript{\textit{l}}}\sum_{\textit{m=1}}^{n}\textit{w}\textsubscript{m}\textsuperscript{1}\textit{f}\textsubscript{m}\textsuperscript{k}(\textit{i},\textit{j})\\
       & = \sum_{\textit{(i,j=1)}}^{\textit{h}/2\textsuperscript{\textit{l}}}\textit{C}\textsubscript{1}\textsuperscript{k}(\textit{i},\textit{j})
  \end{split}
\end{equation}
where \textit{C}\textsubscript{1}\textsuperscript{k}(\textit{i},\textit{j}) is the Class Activation Map for (\textit{i},\textit{j}) spatial location in \textit{k}-th sub-CNN model for predicted class 1. Since the size of input \textit{x} is \textit{h} x \textit{h}, resizing \textit{C}\textsubscript{1}\textsuperscript{k}(\textit{i},\textit{j}) by \textit{h}/$\sqrt{n}$ gives the same size as input \textit{x} and we can define it as \textit{C}\textsubscript{1}\textsuperscript{k}(\textit{x}). From the equation \ref{(equ2.5)}, it can be said that \textit{C}\textsubscript{1}\textsuperscript{k}(\textit{i},\textit{j}) indicates the importance of the activation at spatial location (\textit{i},\textit{j}) leading to the classification to predicted class 1 in \textit{k}-th sub-CNN model. Likewise, \textit{C}\textsubscript{1}\textsuperscript{k}(\textit{x}) represents which pixels of input \textit{x} played an important role in classifying input \textit{x} as a predicted class 1 in \textit{k}-th sub-CNN model. Based on the equations described so far, \textit{C}\textsubscript{0}\textsuperscript{k}(\textit{x}) can be defined as the CAM for predicted class 0 in \textit{k}-th sub-CNN model for given input \textit{x}.

So far we have explained the CAM generation process for input \textit{x} at each sub-CNN model. As a result, input \textit{x} generates two types of CAMs, \textit{C}\textsubscript{0}\textsuperscript{k}(\textit{x}) and \textit{C}\textsubscript{1}\textsuperscript{k}(\textit{x}), in \textit{k}-th sub-CNN model. The next thing to define is combining these \textit{C}\textsubscript{0}\textsuperscript{k}(\textit{x}) and \textit{C}\textsubscript{1}\textsuperscript{k}(\textit{x}) into unified feature for input \textit{x} for aggregated predicted class \textit{P}(\textit{x}) from primitive classification. This unified feature can be seen as Region of Interest (ROI) and defined as \textit{R}(\textit{x}). When generating \textit{R}(\textit{x}), we need to consider that the more distant \textit{P}(\textit{x}) and \textit{k} are, the lower the effect of \textit{C}\textsubscript{0}\textsuperscript{k}(\textit{x}) and \textit{C}\textsubscript{1}\textsuperscript{k}(\textit{x}). For example, if the predicted age at the facial age estimation problem is 20 years old, it is obvious that the sub-CNN model classified by age 19 has a higher influence than the sub-CNN model classified by age 50. Therefore, we introduce distance metric \textit{D}\textsubscript{P}\textsuperscript{k}(\textit{x}) to quantify this influence of CAM for input \textit{x} in \textit{k}-th sub-CNN model. We can define \textit{D}\textsubscript{P}\textsuperscript{k}(\textit{x}) by directly applying background information of inter-class relation. If the actual class \textit{x} has an ordinal relationship, such as an facial age estimation problem, \textit{D}\textsubscript{P}\textsuperscript{k}(\textit{x}) can be expressed by the following equation.
\begin{equation}\label{(equ2.6)}
\textit{D}\textsubscript{P}\textsuperscript{k}(\textit{x})=
\begin{cases}
	\frac{1}{{\textit{P}(\textit{x})-\textit{k}}+1}, & \textit{k} \leq \textit{P}(\textit{x}) \\
	\frac{1}{{\textit{k} - \textit{P}(\textit{x})}}, & \textit{k} > \textit{P}(\textit{x})\\
\end{cases}
\end{equation}
where \textit{P}(\textit{x}) $\in$ \{1, 2, ..., \textit{N}-2\}. If \textit{P}(\textit{x}) is 0 or \textit{N}-1, \textit{D}\textsubscript{P}\textsuperscript{k}(\textit{x}) is not needed because \textit{R}(\textit{x}) is defined differently. Combining \textit{C}\textsubscript{0}\textsuperscript{k}(\textit{x}) and \textit{C}\textsubscript{1}\textsuperscript{k}(\textit{x}) with \textit{D}\textsubscript{P}\textsuperscript{k}(\textit{x}), \textit{M}(\textit{x}) can be defined as follows.
\begin{equation}\label{(equ2.7)}
\textit{R}(\textit{x})=
\begin{cases}
	\sum_{\textit{k} \leq \textit{P}(\textit{x})}\textit{D}\textsubscript{P}\textsuperscript{k}(\textit{x})\textit{C}\textsubscript{0}\textsuperscript{k}(\textit{x}) + 
	\sum_{k > \textit{P}(\textit{x})}\textit{D}\textsubscript{P}\textsuperscript{k}(\textit{x})\textit{C}\textsubscript{1}\textsuperscript{k}(\textit{x}), & \textit{P}(\textit{x}) \in \{1, ..., \textit{N}-2\}\\
	\textit{C}\textsubscript{0}\textsuperscript{1}(\textit{x}), & \textit{P}(\textit{x}) = 0\\
	\textit{C}\textsubscript{1}\textsuperscript{\textit{N}-1}(\textit{x}), & \textit{P}(\textit{x}) = \textit{N}-1\\
\end{cases}
\end{equation}

From the equation \ref{(equ2.7)}, when \textit{k} is less than or equal to \textit{P}(\textit{x}), we multiply the distance metric \textit{D}\textsubscript{P}\textsuperscript{k}(\textit{x}) by the \textit{C}\textsubscript{0}\textsuperscript{k}(\textit{x}) of the \textit{k}-th sub-CNN model. Otherwise, we multiply the \textit{D}\textsubscript{P}\textsuperscript{k}(\textit{x}) with \textit{C}\textsubscript{1}\textsuperscript{k}(\textit{x}). This part can be reversed according to the definition of the user, but from our experimental results, it was better to define it as above. A more intuitive reason is as follows. When \textit{P}(\textit{x}) is aggregated with \textit{p}\textsubscript{k}(\textit{x}), \textit{p}\textsubscript{k}(\textit{x}) is likely to be 1 in the \textit{k}-th sub-CNN model where k is less than or equal to \textit{P}(\textit{x}). For facial age estimation example, if the predicted age is 20 years old, then it is likely that the sub-CNN model classified by age 15 is likely to have output 1 and the model by age 30 is likely to have output 0. In other words, it can be assumed that the abstract representation of \textit{C}\textsubscript{1}\textsuperscript{k}(\textit{x}) is already contained in \textit{P}(\textit{x})  if \textit{k} is less than or equal to \textit{P}(\textit{x}). Therefore, if we create a \textit{R}(\textit{x}) by aggregating the opposite class CAMs, it is presumed that final classification process can be trained with various information which is more likely to correct error of \textit{P}(\textit{x}) with higher probability. We experiment on both combinations and compare the results later in the evaluation. In addition, when \textit{P}(\textit{x}) is 0 or \textit{N}-1, only the CAM from the first or the \textit{N}-1th sub-CNN model are \textit{R}(\textit{x}) without considering the other sub-CNN models. This is because Ranking-CNN performs one-to-all classification for classes at both ends. That is, when \textit{P}(\textit{x}) is 0, the first sub-CNN model can be thought of as a model that directly classifies \textit{P}(\textit{x}) equals 0 and vice versa in case of \textit{P}(\textit{x}) equals \textit{N}-1. Therefore, when \textit{P}(\textit{x}) is 0, it is reasonable to set \textit{R}(\textit{x}) directly with \textit{C}\textsubscript{0}\textsuperscript{1}(\textit{x}) and \textit{C}\textsubscript{1}\textsuperscript{\textit{N}-1}(\textit{x}) when \textit{P}(\textit{x}) is \textit{N}-1. 

The ROI extraction step can be summarized as generating \textit{R}(\textit{x}) for the arbitrary input \textit{x} from \textit{P}(\textit{x}) and weights of sub-CNN models in the primitive classification and passing it to the final classification step. Algorithm \ref{algo:roi_ext} provide the entire process of ROI extraction step.

%% ROI Extraction Algorithm
\begin{algorithm}
\caption{ROI Extraction}\label{algo:roi_ext}
\begin{algorithmic}[1]
\Procedure{CAM Generation Procedure}{}
\For{$ k=1\hspace{2mm}to\hspace{2mm}N-1 $}
\State $ \textit{C}\textsubscript{0}\textsuperscript{k}(\textit{x}) \gets \textit{k-th sub-CNN} $
\State $ \textit{C}\textsubscript{1}\textsuperscript{k}(\textit{x}) \gets \textit{k-th sub-CNN} $
\EndFor
\EndProcedure
\Procedure{ROI Generation Procedure}{}
\State $ P(x) \gets \hspace{2mm}\textit{Prediction Procedure in}\hspace{2mm}\textbf{Algorithm1} $
\If {$ P(x)=0 $}
\State $ R(x) \gets \textit{C}\textsubscript{0}\textsuperscript{1}(\textit{x})$
\ElsIf {$ P(x)=N-1 $}
\State $ R(x) \gets \textit{C}\textsubscript{1}\textsuperscript{N-1}(\textit{x})$
\Else
\State $ R(x) \gets \sum_{\textit{k} \leq \textit{P}(\textit{x})}\textit{D}\textsubscript{P}\textsuperscript{k}(\textit{x})\textit{C}\textsubscript{0}\textsuperscript{k}(\textit{x}) + 
	\sum_{k > \textit{P}(\textit{x})}\textit{D}\textsubscript{P}\textsuperscript{k}(\textit{x})\textit{C}\textsubscript{1}\textsuperscript{k}(\textit{x})$
\EndIf
\EndProcedure
\end{algorithmic}
\end{algorithm}
%-------------------------------------------------------------------------
\subsection{Final Classification}
The role of the final classification step is to combine the \textit{R}(\textit{x}) received from the ROI extraction step with the arbitrary input \textit{x} $\in$ \textit{X} to generate a new input \textit{x'} $\in$ \textit{X'} and perform strict training for final prediction. Algorithm \ref{algo:fin_cla} represents the overall process of final classification step from input \textit{x'} generation to final prediction.
%Final Classification 알고리즘
\begin{algorithm}[]
\caption{Final Classification}\label{algo:fin_cla}
\begin{algorithmic}[1]
\Procedure{Generate Input Procedure}{}
\State $ R(x) \gets \hspace{2mm}\textit{ROI Generation Procedure in}\hspace{2mm}\textbf{Algorithm2} $
\For{$ x\hspace{2mm}\in\hspace{2mm}X $}
\State $ x'\hspace{2mm}\gets\hspace{2mm}x+R(x)$
\EndFor
\State $ x'\hspace{2mm}\in\hspace{2mm}X'$
\EndProcedure
\Procedure{Final Training Procedure}{}
\For{$ k'=1\hspace{2mm}to\hspace{2mm}N-1 $}
\State $ \textbf{initialize}\hspace{2mm}\textit{k'-th sub-CNN} $
\EndFor
\BState \emph{top}:
\For{$ k'=1\hspace{2mm}to\hspace{2mm}N-1 $}
\State $\textit{X'}\textsuperscript{0}\textsubscript{k'} = \{(\textit{x'}, 0) | \textit{y} < k'\}$
\State $\textit{X'}\textsuperscript{1}\textsubscript{k'} = \{(\textit{x'}, 1) | \textit{y} \geq k'\}$
\State $ \textbf{fine-tune}\hspace{2mm}\textit{k'-th sub-CNN} $
\EndFor
\If {$ \textit{not converged} $}
\State $ \textbf{goto}\hspace{2mm}top $
\EndIf
\EndProcedure
\Procedure{Final Prediction Procedure}{}
\For{$ k'=1\hspace{2mm}to\hspace{2mm}N-1 $}
\State $ \textit{p\textsuperscript{k'}(x')} \gets \textit{k'-th sub-CNN} $
\EndFor
\State $ \textit{P(x')} \gets \sum_{\textit{k'}=1}^{\textit{N}-1}\textit{p\textsuperscript{k'}(x')} $
\EndProcedure
\end{algorithmic}
\end{algorithm}

There are several ways to combine input \textit{x} and \textit{R}(\textit{x}), but we define input \textit{x'} with additional channels for \textit{R}(\textit{x}) to preserve the information of original x. That is, when input \textit{x} is the size of \textit{h} x \textit{h} x 3, then the new input \textit{x'} is the size of \textit{h} x \textit{h} x 4 with the \textit{R}(\textit{x}) of size \textit{h} x \textit{h} appended. The advantage of this method is that even if the input x is augmented during the training, the spatial information of \textit{R}(\textit{x}) can be maintained by applying same augmentation policy. In other words, if input \textit{x'} is shifted, rotated, and resized, both input \textit{x} and \textit{R}(\textit{x}) are applied in the same way. The process of classifying the input \textit{x'} is similar to the primitive classification, but there is no need to output CAM, so adding fully-connected layer after the last convolutional layer is no longer restricted. Once the training is finished, evaluation procedure is done with the test-set that was separated from the beginning. As we mentioned in the previous section, \textit{R}(\textit{x}) for the \textit{P}(\textit{x}) from the previous two steps should be combined with the test-set to be classified into the correct class.
%-------------------------------------------------------------------------
\section{TRk-CNN for glaucoma detection}
\label{ch2:trk-glaucoma}
\begin{figure}[!tbh]
	\centering
	\includegraphics[width=\textwidth]{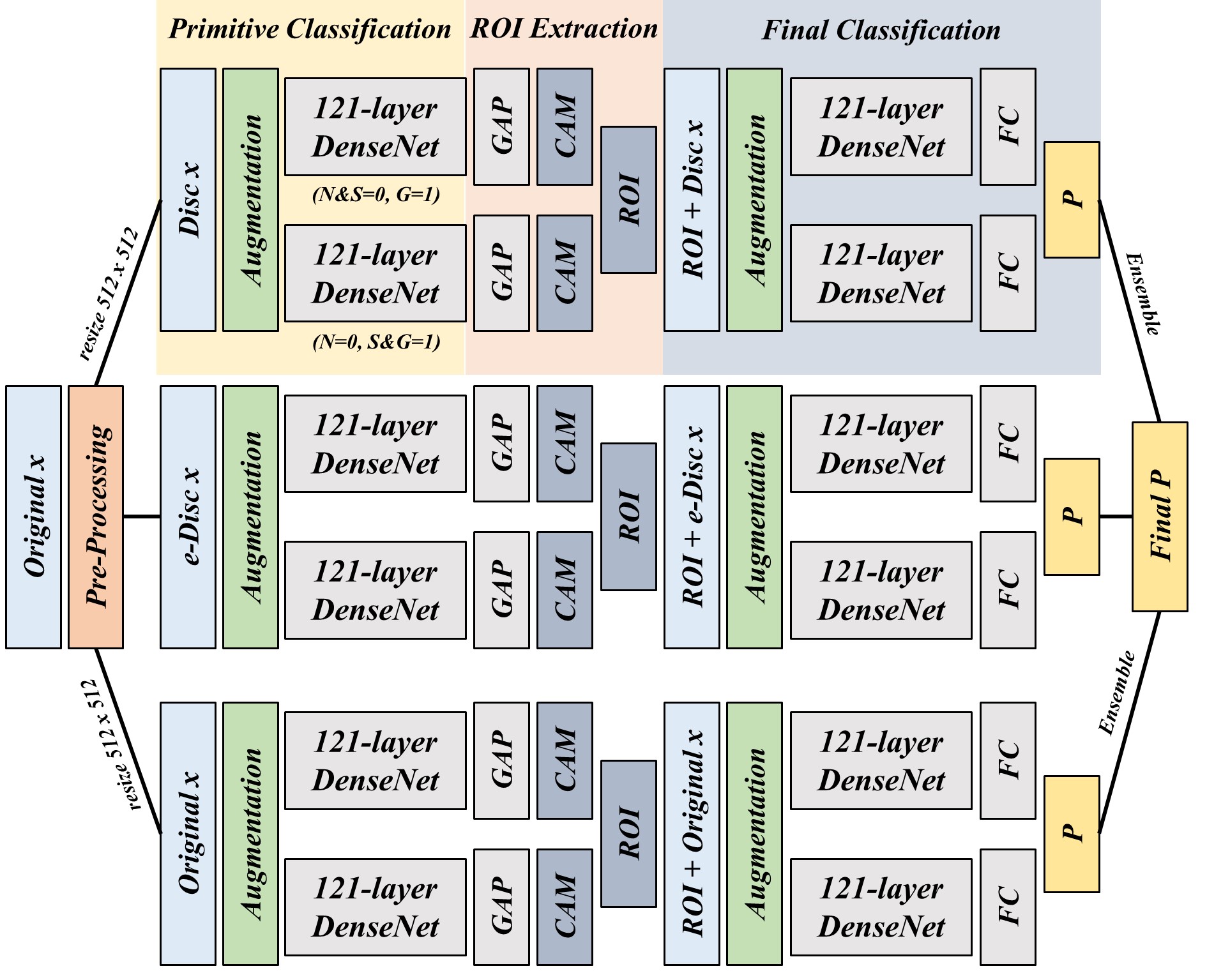}
	\caption{The overall overall process of TRk-CNN for glaucoma detection}
	\label{img:trk_process}
\end{figure}

In this section, we introduce the method of glaucoma detection based on the TRk-CNN. The fundus images we want to classify are labeled as normal, glaucoma suspect, and glaucoma eyes. Since glaucoma suspicious eyes can be seen as an intermediate stage between normal eyes and glaucoma eyes, better performance can be achieved by considering the inter-class relationship with TRk-CNN. The overall process of glaucoma detection is as follows. First, we perform pre-processing on fundus images. Then, the fundus images are augmented to perform primitive classification. Next, the ROI is generated from the predicted value and the weight of the sub-CNN model obtained as results of the primitive classification. Finally, the ROI is combined with the original fundus image to perform the final classification, and the aggregated predicted class is compared with the actual class. Figure \ref{img:trk_process} shows the overall process for classifying normal, glaucoma suspect, and glaucoma eyes with TRk-CNN.
%상자들 나오면서 녹내장 분류의 전체적인 조직도 보여주는 그림

\subsection{Pre-processing}
\begin{figure}[!tbh]
	\centering
	\includegraphics[width=\textwidth]{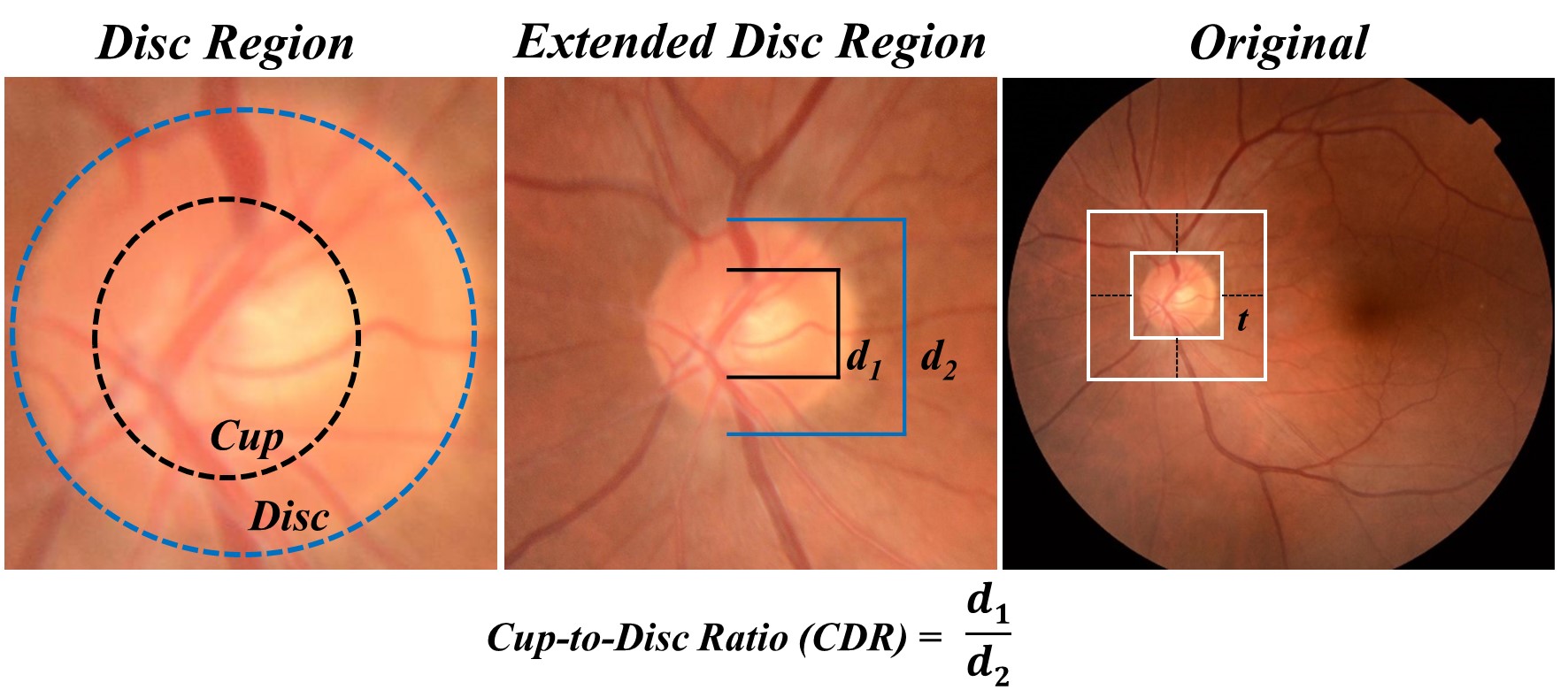}
	\caption{Three regions of pre-processing results}
	\label{img:pre_proc}
\end{figure}

Although the resolution of the fundus image is very high, the area that plays an important role in the diagnosis of glaucoma is the disc/cup region. This is because cup-to-disc ratio (CDR) is one of the main criteria for discriminating glaucoma suspicious eyes. Therefore, in the pre-processing stage, we manually extract the disc/cup region of the fundus images. The optimized model will apply TRk-CNN models to the original fundus image, disc region image, and extended disc (e-disc) region image and then ensemble the results of the three models. The extended disc region is a region where the same range of pixels (\textit{t}) is added to the top, bottom, left, and right sides of the disc region that we manually extracted. Therefore, the extended disc region can be regarded as the intermediate image between the disc region and the original image. There are several previous studies that automatically segments the disc/cup region with machine learning approaches. However in this paper, we believe it is sufficient to draw it manually because the area we are interested in is a square box that contains disc/cup, not the exact pixel-by-pixel disc/cup region. And although our the evaluation results show that applying TRk-CNN to the disc region has the highest performance, it is not much different from the results of the other two images. Figure \ref{img:pre_proc} shows images of the three different regions from the same fundus images obtained as a result of pre-processing.

\subsection{Data augmentation}
\begin{figure}[!tbh]
	\centering
	\includegraphics[width=\textwidth]{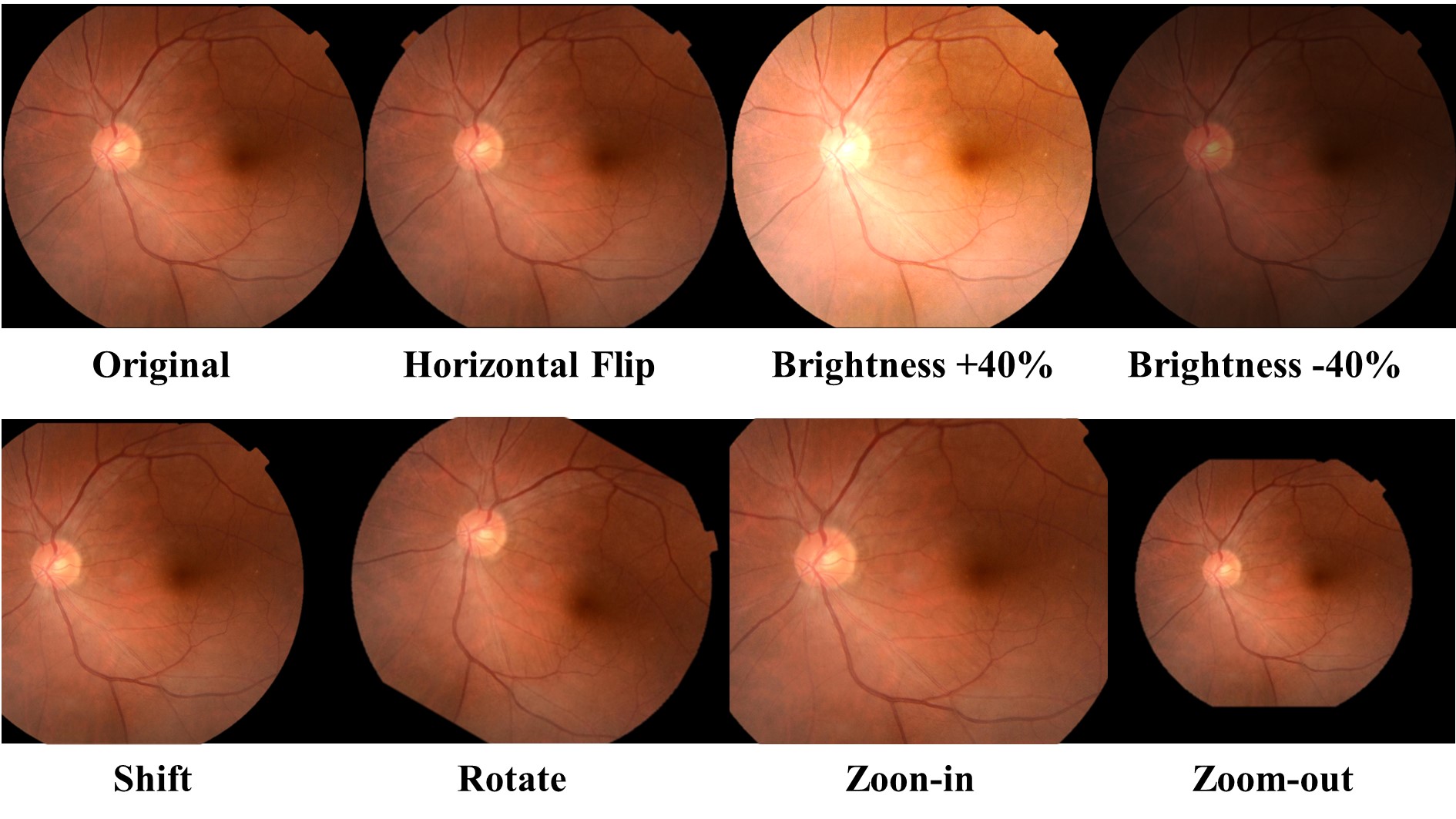}
	\caption{Example images from data augmentation}
	\label{img:augment}
\end{figure}

Since our data set consists of about 1,000 fundus images, without augmentation the model will fall into overfitting problem shortly and it will be hard to expect reasonable performance for validation and test set. Fortunately, fundus images are not as varied as the general dataset such as ImageNet or Cifar10. In other words, normal, glaucoma suspicious, and glaucoma eyes are classified from fundus images with relatively similar class distribution, compared to a general image dataset with a heterogeneous class distribution. Therefore, even with a thousand number of images, the proper application of augmentation can yield acceptable classification accuracy. Our image augmentation policy is as follows. First, we zoom-in and zoom-out an image at a random ratio within $\pm$20\%. And the height and width of the image are shifted at a random ratio within $\pm$20\% of image size \textit{h} x \textit{h}. Also, the image flips horizontally with a random probability, which has the effect of augmenting the right eye into the left eye and vice versa. Next, since the fundus image may have a different eye orientation depending on the angle of the screening, we rotate the image within $\pm$45\textsuperscript{$\circ$} at random rates. Finally, because the brightness of the fundus image is also different, the brightness is also changed within $\pm$40\% at random rates. Figure \ref{img:augment} shows images when each augmentation policy is applied to a single image at the maximum rate.

\subsection{Primitive Classification}
Starting from the primitive classification, the backbone structure of the CNN model to be used in the following steps is the DenseNet \cite{ch23_densenet} with 121 number of layers. DenseNet extends ResNet's \cite{ch23_resnet} skip-connection concept and is characterized by a densely connected block. Dense connection encourages feature reuse and reduces the number of free parameters, thereby reducing overfitting in a relatively small train-set. Therefore, we judged that DenseNet as backbone CNN model is suitable for our fundus dataset which has smaller train-set than general image dataset such as ImageNet \cite{ch23_imagenet} and Cifar10 \cite{ch2_cifar}. However, the free parameters of DenseNet are still many to optimize with a thousand fundus images. So we started training by taking the weight of pre-trained 121-layer DenseNet in ImageNet and this method is generally called as transfer learning. Of course, because ImageNet images and fundus images are different types of images, we trained the entire weight from the beginning, unlike the general transfer learning method which trains only a few top layers. 

Before starting training, the train-set is transformed according to the augmentation policy. And an input image is resized to 512 x 512 x 3 in all regions including original, disc, and e-disc region. The reason for adjusting the image size to 512 x 512, which is larger than common sizes 224 x 224 or 256 x 256, is because the size of the original fundus image is very large, which has a minimum size of 3500 x 2500. The resized and augmented train-set with the mini-batch size is now passed to the input of the 121-layer DenseNet to start training.

\begin{figure}[!tbh]
	\centering
	\includegraphics[width=\textwidth]{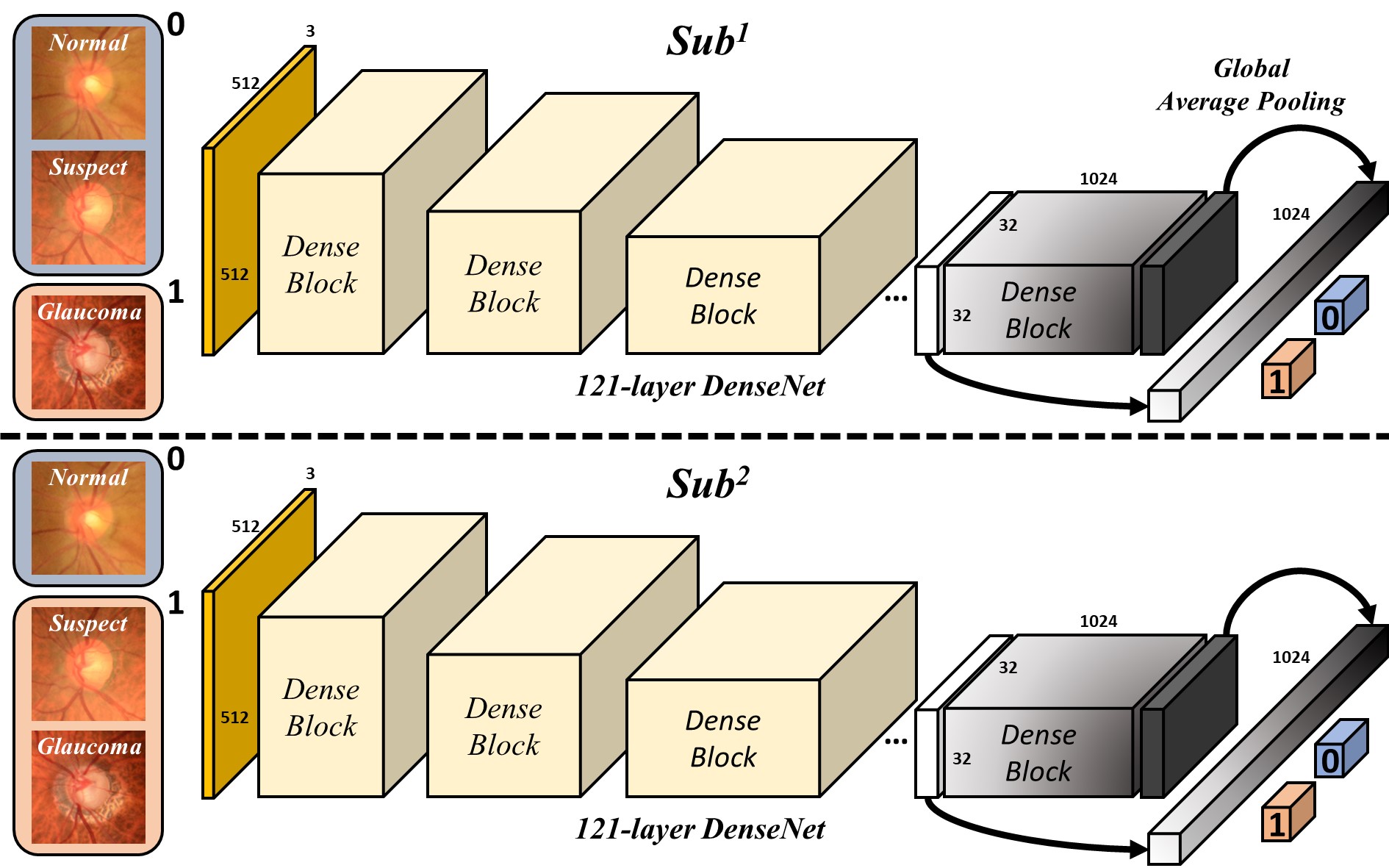}
	\caption{Primitive classification for glaucoma detection}
	\label{img:pri_cla}
\end{figure}

Since our fundus dataset has three classes, two sub-CNN models are required to perform Ranking-CNN in primitive classification. We labeled the actual class of normal eye as 0, glaucoma suspect eye as 1, and glaucoma eye as 2. Of course, the actual class of normal and glaucoma eye may be interchanged, but the existence of a glaucoma suspect eye between them should be maintained to perform Ranking-CNN. Let the 1st sub-CNN model as \textit{Sub}\textsuperscript{1} and the 2nd sub-CNN model as \textit{Sub}\textsuperscript{2}. Then input class of \textit{Sub}\textsuperscript{1} is 0 for normal eye, 1 for glaucoma suspect and glaucoma eyes. Likewise, in \textit{Sub}\textsuperscript{2}, normal and glaucoma suspect eyes become class 0, and glaucoma eye becomes class 1. After the input is passed to each sub-CNN model with the 121-layer DenseNet, the size of a final convolutional layer is 32 x 32 x 1024. Applying the global average pooling results in a layer with a size of 1024, followed by a size 2 softmax layer for binary classification. The optimization parameters of the model will be explained more concretely in the final classification section. As a result, the weight of the model with a minimum loss for the validation set and the aggregated predicted class \textit{P} $\in$ \{0,1,2\} for the input are passed to the ROI extraction step. The aggregated predicted class \textit{P} can be obtained as \textit{P} = \textit{p}\textsubscript{1} + \textit{p}\textsubscript{2}, where \textit{p}\textsubscript{1} $\in$ \{0,1\} is the predicted class of \textit{Sub}\textsuperscript{1} and \textit{p}\textsubscript{2} $\in$ \{0,1\} is the predicted class of \textit{Sub}\textsuperscript{2} for the input. Figure \ref{img:pri_cla} shows the overall process of primitive classification step for glaucoma detection.

\subsection{ROI Extraction}
The purpose of the ROI extraction step is to generate the region of interest \textit{R} based on the \textit{P} and model weight received earlier from the primitive classification. First, the Class Activation Maps for the binary classes of \textit{Sub}\textsuperscript{1} and \textit{Sub}\textsuperscript{2} for the given input are called \textit{Cam}\textsubscript{0}\textsuperscript{1}, \textit{Cam}\textsubscript{1}\textsuperscript{1}, \textit{Cam}\textsubscript{0}\textsuperscript{2}, and \textit{Cam}\textsubscript{1}\textsuperscript{2}, respectively. That is, \textit{Cam}\textsubscript{0}\textsuperscript{1} is the CAM of \textit{Sub}\textsuperscript{1} as class 0 for the given input. Since the size of the input excluding the channel is 512 x 512 and the number of nodes in the GAP is 1024, we obtain the size 512 x 512 CAM by resizing the output by 512/$\sqrt{1024}$ times, which is 16. Based on the equation \ref{(equ2.7)} the ROI \textit{R} for the predicted class \textit{P} can be expressed by the following equation.
\begin{equation}\label{(equ2.8)}
\textit{R}=
\begin{cases}
	\textit{Cam}\textsubscript{0}\textsuperscript{1}, & \textit{P} = 0\\
	\textit{Cam}\textsubscript{0}\textsuperscript{1} +  \textit{Cam}\textsubscript{1}\textsuperscript{2}, & \textit{P} = 1\\
	\textit{Cam}\textsubscript{1}\textsuperscript{2}, & \textit{P} = 2
\end{cases}
\end{equation}

\begin{figure}[!tbh]
	\centering
	\includegraphics[width=\textwidth]{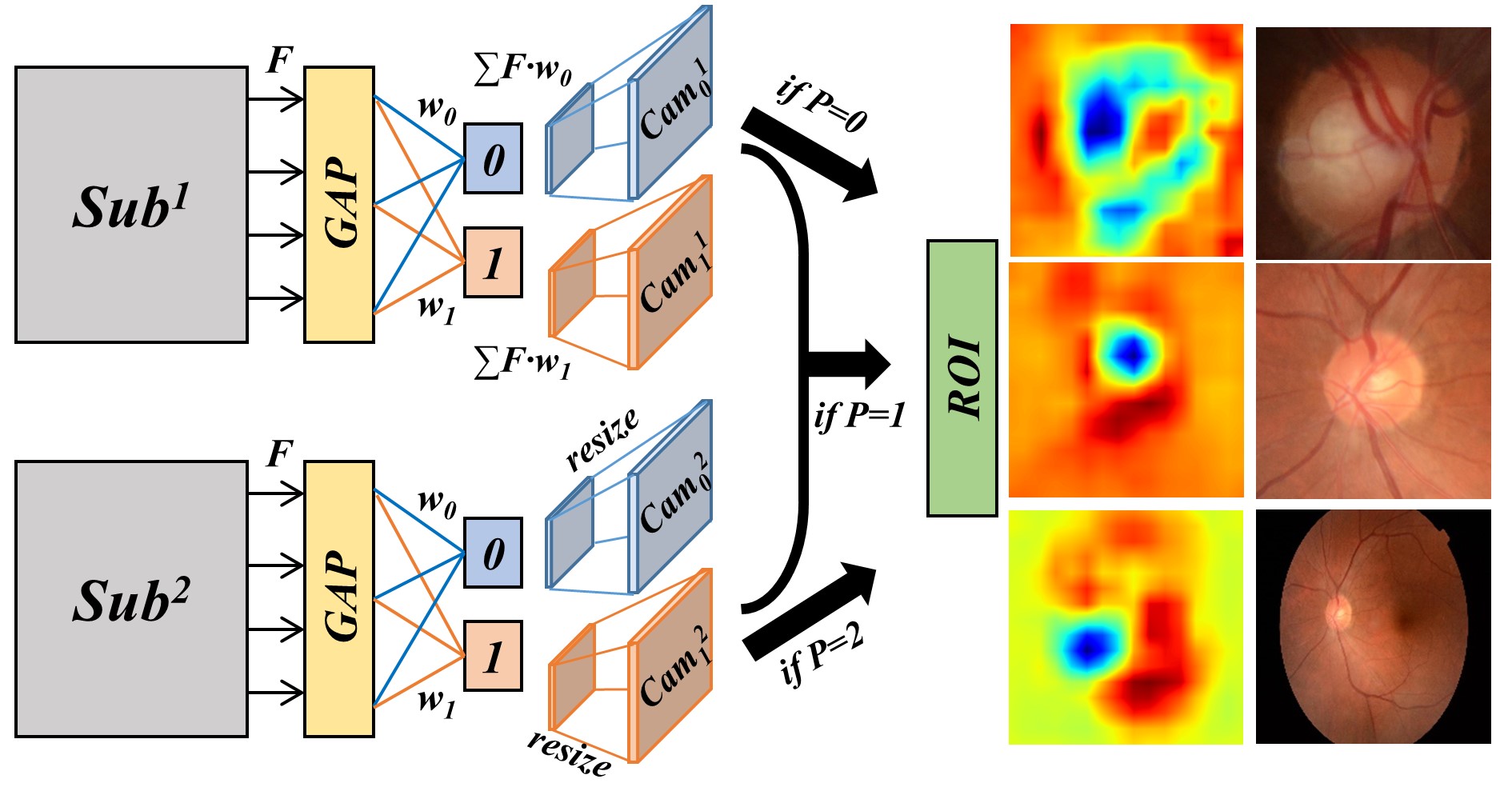}
	\caption{ROI extraction for glaucoma detection}
	\label{img:roi_ext}
\end{figure}

Finally we perform z-score normalization before passing the generated \textit{R} to the final classification step. As we mentioned in section 2.3.2, we have also evaluated \textit{R} =  \textit{Cam}\textsubscript{1}\textsuperscript{1} +  \textit{Cam}\textsubscript{0}\textsuperscript{2} when the \textit{P} is 1 in Result section to compare the performance difference. Figure \ref{img:roi_ext} shows the overall process of ROI extraction step for glaucoma detection.

\subsection{Final Classification}
The final classification begins by concatenating the input with \textit{R} from the ROI extraction step. Since the train-set has a size of 512 x 512 x 3 and \textit{R} has a size of 512 x 512, if we concatenate the two, the size of the new train-set is 512 x 512 x 4. The image augmentation policy is the same as the primitive classification, but for brightness policy, it should only be applied to the original train-set of the new train-set. The reason is that the last channel, which is \textit{R}, should be transformed with the rotation, translation, and zooming policies because \textit{R} represents the spatial characteristics of the given input, but it is not affected by brightness. The newly generated input is classified through Ranking-CNN based on 121-layer DenseNet as well as the primitive classification step. One difference now is that strict training is available by adding fully-connected layers after the GAP layer. From here, we will explain a specific description of the parameters applied to the final classification.

\subsubsection{Loss Function}
In general, categorical cross-entropy loss (\textit{CELoss}) is used for the loss function in classification problems, but in our experience, using categorical cross-entropy (\textit{CE}) alone increases the gap between minimum validation loss and maximum validation accuracy (\textit{Acc}). As will be described later in the evaluation, intuitively, the gap between the softmax output vector and the predicted class vector occurs when argmax function is applied. Therefore, we use a loss function that combines both categorical cross-entropy loss and average accuracy. When we use categorical cross-entropy loss alone in the glaucoma detection problem, we confirmed that it converges at a validation loss of about 0.1 and that the validation accuracy converges to around 0.9. However, since the fluctuation of cross entropy per epoch is greater than the fluctuation of accuracy, we needed to adjust the scale of categorical cross-entropy loss from the final loss. As a result, the categorical cross-entropy loss with accuracy (\textit{CEALoss}) for input \textit{x} = \{\textit{x}\textsubscript{1}, \textit{x}\textsubscript{2}, ... , \textit{x}\textsubscript{b}\} with mini-batch size \textit{b} is as follows.
%-\sum_{i=1}^{c}{y\textsubscript{i}(x)\ln o\textsubscript{i}(x) + (1-y\textsubscript{i}(x))\ln (1-o\textsubscript{i}(x))}
\begin{equation}\label{(equ2.9)}
\begin{gathered}
	CE(x) = 
	-\sum_{i=1}^{c}{\ln s\textsubscript{i}(x)}\\
	CELoss = 
	\frac{1}{b}\sum_{j=1}^{b}CE(x\textsubscript{j})\\
	Acc = \frac{1}{b}\sum_{j=1}^{b}{y(x\textsubscript{j}) \cdot p(x\textsubscript{j})}\\
	CEALoss = 
    1 + \alpha CELoss - Acc
\end{gathered}
\end{equation}
where \textit{c} is the number of classes, \textit{s}\textsubscript{i}(\textit{x}) is the softmax output value for class \textit{i} $\in$ \{1, 2, ..., \textit{c}\}, \textit{y}(\textit{x}) is the one hot encoded vector represents true class for input \textit{x}, \textit{p}(\textit{x}) is the one hot encoded vector represents predicted class for input \textit{x}, and $\alpha$ is coefficient for adjusting the scale of \textit{CELoss} which set to 0.1 in this paper. However, $\alpha$ can be intuitively changed depending on the classification problem. We compared the performance of the \textit{CEALoss} and the \textit{CELoss} in the evaluation, and as a result, the performance of the \textit{CEALoss} was better.

\subsubsection{Activation and Optimizer Functions}
The role of the activation function is to define the output value of kernel weights in the model. In modern CNN models, nonlinear activation is widely used, including rectified linear units (ReLU) \cite{ch23_relu}, leakage rectified linear units (LReLU) \cite{ch2_lrelu}, and exponential linear units (ELU) \cite{ch2_elu}. As we experimentally confirmed, we have applied the most commonly used ReLU because the three activation functions were not significantly different in performance.

The role of the optimizer function is to minimize the loss function through the stochastic gradient descent approach with learning rate. There are several well-known optimizer functions such as Adam \cite{ch23_adam}, Adagrad \cite{ch2_adagrad}, and Adadelta \cite{ch2_adadelta}.
In general, Adam function converges faster than other functions. Therefore, we also used Adam for optimizer function and the initial learning rate was set to 0.0001. In addition, we reduced the learning rate by half if the validation loss does not improve for the last 10 epochs.

\subsubsection{Regularization}
Regularization is a method to reduce overfitting during the training phase. Overfitting is a problem especially when the size of the train-set is small and the free parameter of the model is large like our glaucoma detection problem. Image augmentation is also a regularization technique, which is not directly applied to the model, so it is described after the pre-processing section. Typical regularization methods are using L1 and L2 norm, however, it is common to apply Dropout \cite{ch2_dropout} and Batch Normalization \cite{ch23_bn} in recent CNN models. In deep learning, when a layer is deepened, a small parameter change in the previous layer can have a large influence on the input distribution of the later layer. This phenomenon is referred to as internal co-variate shift. Batch normalization has been proposed to reduce this internal co-variate shift, and the mean and variance of input batches are calculated, normalized, and then scaled and shifted. The location of Batch Normalization is usually applied just before the activation function and after the convolution layer. The 121-layer DenseNet \cite{ch23_densenet} we used uses Batch Normalization by default, and it is also applied to the last two fully-connected layers.

Another popular regularization technique is Dropout, which stochastically participates in nodes in the same layer, reducing dependency between layers to prevent overfitting. In the training phase, Dropout intentionally excludes some networks, so the model can achieve the voting effect through a combination of partial models. In recent, however, only Batch Normalization is applied to the convolution layers, and Dropout has been selectively added to the fully-connected layer. We also apply Dropout of 0.5 probability to only the last two fully-connected layers.

\begin{figure}[!tbh]
	\centering
	\includegraphics[width=\textwidth]{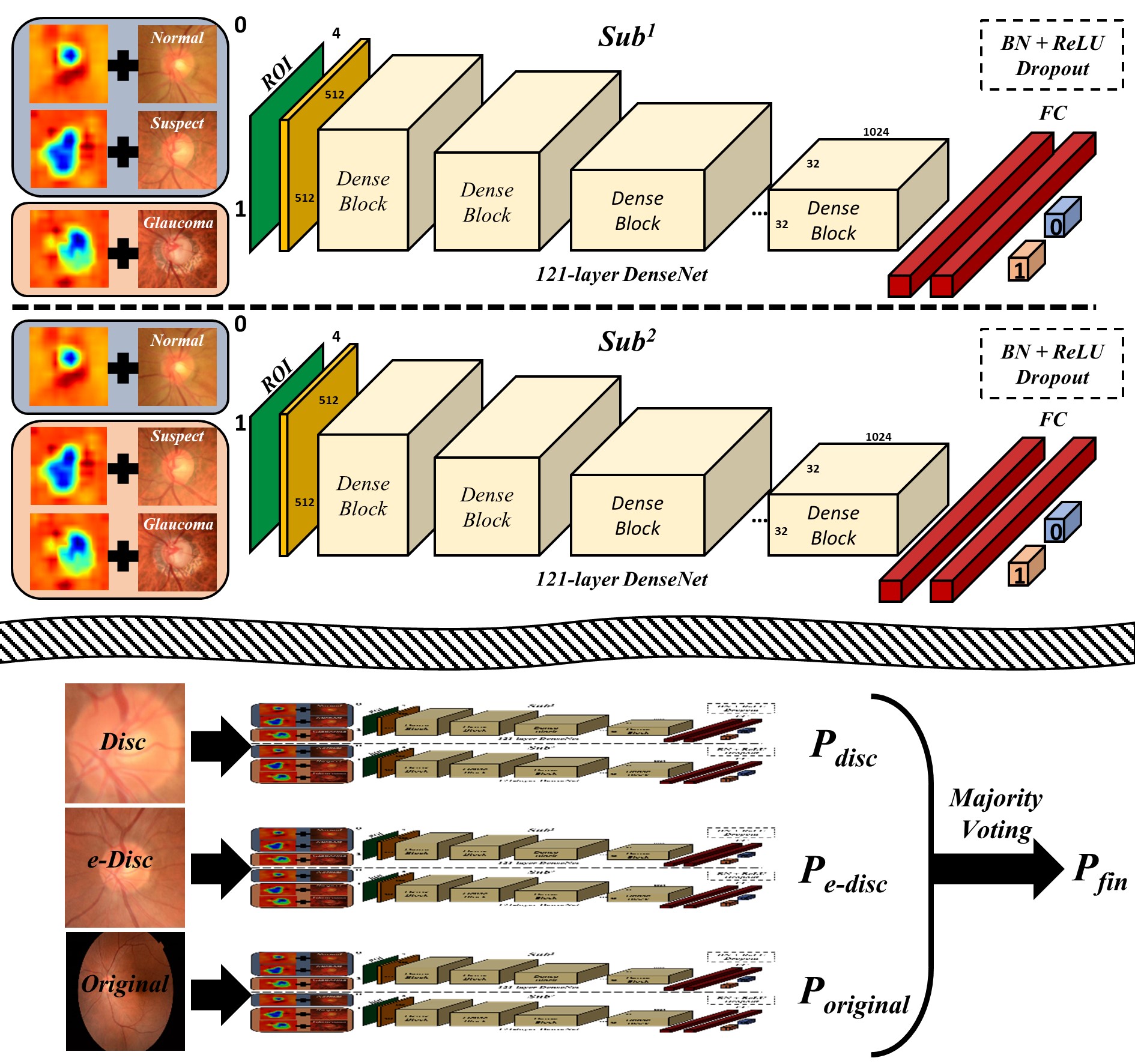}
	\caption{Final classification for glaucoma detection}
	\label{img:fin_cla}
\end{figure}

Finally, an ensemble of several models can be regarded as regularization from the viewpoint of machine learning. In this paper, we use the ensemble method of voting the three prediction results of the trained models from different image regions including original, disc, and e-disc regions. Figure \ref{img:fin_cla} shows the concrete process of the final classification together with optimization parameters.

%-------------------------------------------------------------------------
\section{Results}
\label{ch2:results}
\subsection{Data acquisition}
%데이터를 얻은 과정 (몇 명의 환자, 어떤 환자 등) 그리고 정상/의심/녹내장을 label한 기준
This study included 1022 fundus images from 301 consecutive patients (582 eyes) who underwent fundus imaging with a non-mydriatic fundus camera (TRC-NW8; Topcon, Oakland, NJ, USA), at Korea University Ansan Hospital between January 2016 and August 2017. During the study period, patient electronic medical records and fundus imaging were reviewed to determine the presence of glaucoma by the glaucoma specialist. Based on fundus imaging and electronic medical records, 1022 fundus images were divided into three categories; normal, glaucoma suspect (suspicious), and glaucoma. Fundus images were classified as a glaucoma suspect when a vertical cup-to-disc ratio (CDR) is greater than 0.7 or the peripapillary retinal nerve fiber layer (RNFL) has a characteristic thinning (the presence of RNFL defect) but there is no glaucomatous visual field loss. Fundus images were classified as glaucoma when there is a RNFL defect or visual field loss with a corresponding glaucomatous optic disc change. When fundus images do not correspond to the two mentioned categories above, they were classified as normal. This study adhered to the Declaration of Helsinki and approval for retrospective review of clinical records was obtained from Korea University Ansan Hospital Institutional Review Board (2017AS0036). The patient information was completely anonymized and de-identified prior to analysis. Of the 301 patients, 138 (45.8\%) were men and 163 were women. The mean age ($\pm$ SD) was 59.7 ($\pm$ 15.4) years (range, 19-92 years). There were 291 right eyes (50.0\%) and 291 left eyes. However, 992 images were used as the fundus image dataset of this study because 30 images of the wrong file format were excluded. Of the 922 fundus images, 403 (40.6\%) were normal, 208 (21.0\%) were glaucoma suspect, and 381 (38.4\%) were glaucoma eyes.
%('Train: Total(', 674, ')', {0: 272, 1: 142, 2: 260})
%('Test: Total(', 119, ')', {0: 56, 1: 25, 2: 38})
From the total number of 992 fundus images, 793 images (80\%) were randomly split into train-set and 199 images into (20\%) test-set with the similar class distribution. 199 test-set images consisted of 75 normal images, 41 glaucoma suspect images, and 83 glaucoma images. Validation-set consists of 119 images which correspond to 15\% of the train-set, and also the class distribution is similar. As a result, 674 train-set images consisted of 272 normal images, 142 glaucoma suspect images, and 260 glaucoma images. Likewise, of the 119 validation-set images, 56 images are normal, 25 images are glaucoma suspect, and 38 images are glaucoma eyes. 

\subsection{Evaluation setup}
The software and hardware environment for the evaluation are as follows. We tested on a 64GB server with two NVIDIA Titan X GPUs and an Intel Core i7-6700K CPU. The operating system is Ubuntu 16.04, and the development of the CNN model uses Python-based machine learning libraries including Keras \cite{ch23_keras}, Scikit-learn \cite{ch23_sklearn}, and TensorFlow \cite{ch23_tensorflow}.

\begin{figure}[!tbh]
	\centering
	\includegraphics[width=\textwidth]{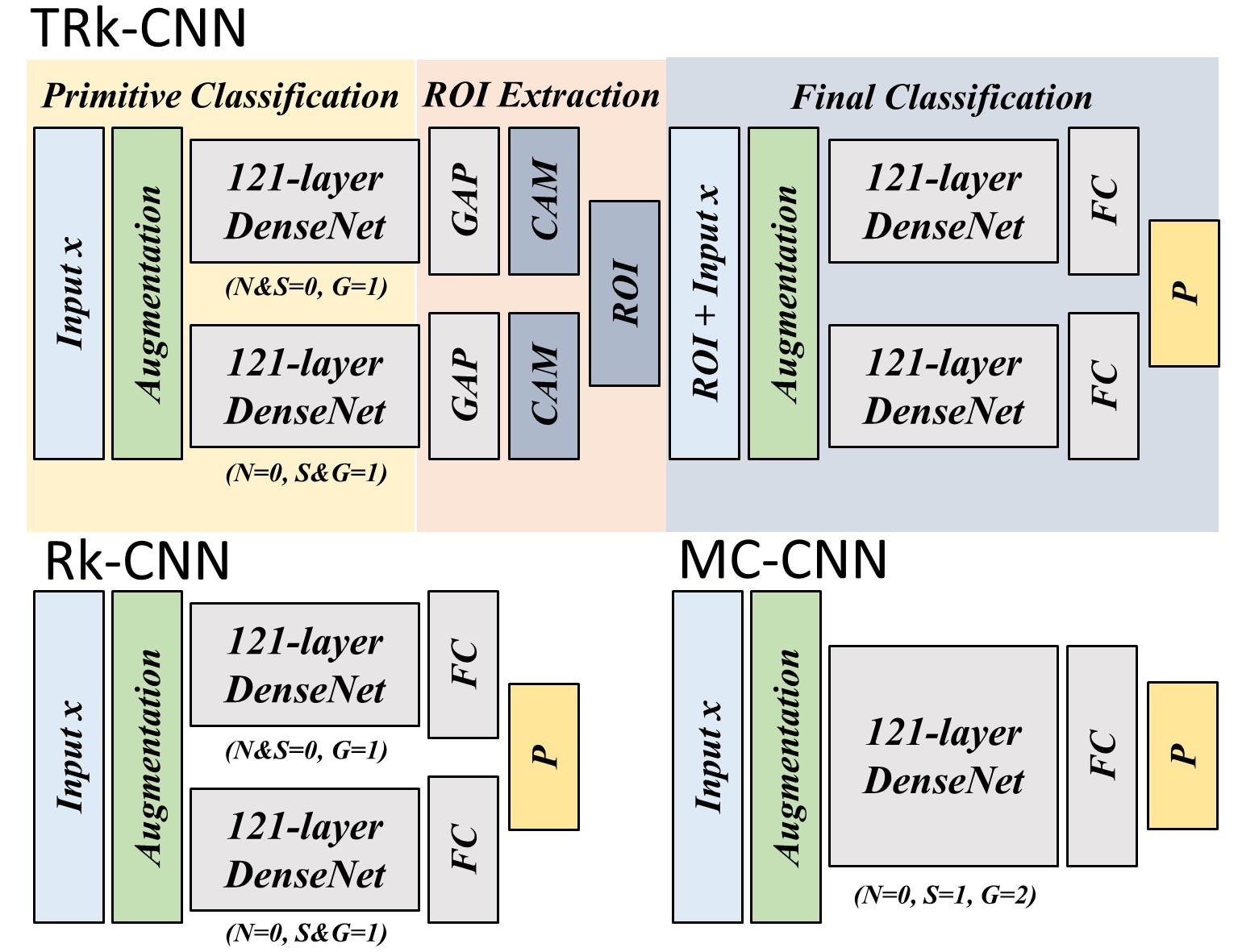}
	\caption{Structural differences between TRk-CNN, Rk-CNN, and MC-CNN}
	\label{img:eval_models}
\end{figure}

We conducted the evaluation from two perspectives. The first is to compare TRk-CNN with Ranking-CNN (Rk-CNN) and multi-class CNN (MC-CNN) under the same conditions. Here, the same condition means that the region of fundus images and the structure of the model are the same. First, the fundus image with a disc region is only used because a disc region shows the best performance among the three regions. Experimental results in three regions are shown by applying TRk-CNN. The same augmentation policy was then applied to train-set images, which is described in detail in the previous section. Rk-CNN and MC-CNN have a 121-layer DenseNet as a basic structure, and two fully-connected layers are added after the last convolutional Layer. In other words, the structure of Rk-CNN and MC-CNN is the same as that of TRK-CNN's final classification step which is shown in Figure \ref{img:eval_models}. The only difference is that MC-CNN classifies normal, glaucoma suspect, and glaucoma at once, so the last prediction layer consists of three nodes. Figure \ref{img:eval_models} outlines the structural differences between TRk-CNN, Rk-CNN, and MC-CNN. Finally, these three models are trained for 100 epochs with the Adam optimizer function with an initial learning rate set to 0.0001, and the learning rate is halved if there is no improvement in validation loss over 10 epochs. The loss function used for comparison is the \textit{CELoss}, not the \textit{CEALoss}. \textit{CEALoss} is used in the optimal TRk-CNN model for glaucoma detection.

The second is to evaluate the different TRk-CNN models, which is optimized for glaucoma detection. We applied the \textit{CEALoss} described in the previous section, and the final prediction was obtained by an ensemble of the three models trained in the original, disc, and e-disc regions of fundus image. The results of the binary classification of each sub-CNN model from the optimized TRk-CNN model are also compared with the performance of the existing literature. Other training parameters and model structure are the same as the three models described above.
%-------------------------------------------------------------------------
\subsection{Evaluation metrics}
The evaluation of the glaucoma classification was based on the following four metrics: average accuracy (\textit{Acc}), specificity (\textit{Sp}), sensitivity (\textit{Se}\textsuperscript{S} for glaucoma suspect, \textit{Se}\textsuperscript{G} for glaucoma), precision (\textit{Pr}\textsuperscript{S}, \textit{Pr}\textsuperscript{G}), and F1 score (\textit{F1}\textsuperscript{S}, \textit{F1}\textsuperscript{G}). Average accuracy means a correctly predicted percentage of the total data. Specificity, also known as the true negative rate, measures the percentage of negatives that are correctly identified as normal. Sensitivity, also known as the true positive rate or recall, measures the percentage of positives that are correctly identified as glaucoma suspect or glaucoma. Precision measures the percentage of positives that are predicted as glaucoma suspect or glaucoma. F1 score is a harmonic mean of sensitivity and precision. These metrics are defined with the following four terminologies.
\begin{itemize}
\setlength\itemsep{0em}
\item True Positive(\textit{TP}): The number of fundus images correctly identified as glaucoma suspect or glaucoma.
\item False Positive(\textit{FP}): The number of fundus images incorrectly identified as glaucoma suspect or glaucoma.
\item True Negative(\textit{TN}): The number of fundus images correctly identified as normal
\item False Negative(\textit{FN}): The number of fundus images incorrectly identified as normal
\end{itemize}
\begin{equation}\label{(2.11)}
\begin{gathered}
    Accuracy(Acc) = \frac{TP + TN}{TP + TN + FP + FN} \times 100 (\%)\\
	Specificity(Sp) = \frac{TN}{FP + TN} \times 100 (\%)\\
	Sensitivity(Se) = \frac{TP}{TP + FN} \times 100 (\%)\\
	Precision(Pr) = \frac{TP}{TP + FP} \times 100 (\%)\\
	F1-score(F1) = \frac{2\times Pr \times Se}{Pr + Se}
\end{gathered}
\end{equation}

\subsection{Evaluation results of TRk-CNN, Rk-CNN, and MC-CNN}
Table \ref{tab:result_three} shows the evaluation results of TRk-CNN, Rk-CNN, and MC-CNN models experimented under the same condition explained in the previous section.
\begin{table}[!tbh]
  \begin{center}
    \scalebox{0.8}{
    \begin{tabular}{|l|c|c|c|c|c|c|c|c|}
     \hline
      Method & \textit{Acc}(\%) & \textit{Sp}(\%) & \textit{Se}\textsuperscript{S}(\%) & \textit{Se}\textsuperscript{G}(\%) & \textit{Pr}\textsuperscript{S}(\%) & \textit{Pr}\textsuperscript{G}(\%) & \textit{F1}\textsuperscript{S}(\%) & \textit{F1}\textsuperscript{G}(\%)\\
     \hline\hline
     TRk-CNN & \textbf{88.94} & \textbf{89.33} & \textbf{85.37} & \textbf{90.36} & 74.47 & 94.94 & \textbf{79.55} & \textbf{92.59}\\
     Rk-CNN & 84.92 & 85.33 & 85.37 & 84.34 & 60.34 & \textbf{100.0} & 70.71 & 91.50\\
     MC-CNN & 83.42 & 85.33 & 68.29 & 89.16 & \textbf{75.68} & 85.06 & 71.79 & 87.06\\
     \hline\hline
     MC-CNN\textsuperscript{1} & 91.46 & 89.33 & \multicolumn{2}{c|}{92.74} & \multicolumn{2}{c|}{93.50} & \multicolumn{2}{c|}{93.12}\\
     MC-CNN\textsuperscript{2} & 92.96 & 97.41 & \multicolumn{2}{c|}{86.75} & \multicolumn{2}{c|}{96.00} & \multicolumn{2}{c|}{91.14}\\
     \hline
     \end{tabular}}
  \end{center}
  \caption{Comparison results between TRk-CNN, Rk-CNN, and MC-CNN}
  \label{tab:result_three}
\end{table}

Overall, TRk-CNN showed higher performance in all metrics except precision. In terms of accuracy, TRk-CNN achieved 88.94\%, which is 4.02\% higher than Rk-CNN and 5.52\% higher than MC-CNN. From the specificity perspective, TRk-CNN was the highest at 89.33\%, which is 4\% higher than Rk-CNN and MC-CNN. The sensitivities of glaucoma suspect for TRk-CNN and Rk-CNN were 85.37\%, which is 17.08\% higher than MC-CNN. The precision of glaucoma suspect for TRk-CNN achieved 74.47\%, which is 14.13\% higher than Rk-CNN and 1.21\% lower than MC-CNN. Since sensitivity and precision have a trade-off relation, it is better to consider F1 score together. In terms of F1-score for glaucoma suspect, TRk-CNN was the highest at 79.55\% which is 7.84\% higher than Rk-CNN and 6.76\% higher than MC-CNN. The sensitivity of glaucoma for TRk-CNN was 90.36\% while Rk-CNN was 84.34\% and MC-CNN was 89.16\%. The precision of glaucoma for TRk-CNN achieved 94.94\% while Rk-CNN was 100.00\% and MC-CNN was 85.06\%. Finally, F1-score of glaucoma for TRk-CNN was the highest at 92.59\% which is 1.09\% higher than Rk-CNN and 5.53\% higher than MC-CNN.

The reason why MC-CNN has the lowest overall performance is that MC-CNN assumes three classes as independent classes without considering the inter-class relationship. For a more precise description, we evaluated MC-CNN to perform binary classification. MC-CNN\textsuperscript{1} classifies normal eye as 0, glaucoma suspicion and glaucoma eyes as 1. Likewise, MC-CNN\textsuperscript{2} classifies normal and glaucoma suspect eyes as 0, glaucoma eye as 1. From Table \ref{tab:result_three}, we can observe that the overall performance is improved despite the same structure as MC-CNN. This shows that our classification problem is a difficult problem compared to the binary classification problem that classifies the normal eyes and glaucoma eyes in the previous studies. We will show the results in comparison with the previous studies in the following section.

\begin{figure}[!tbh]
	\centering
	\includegraphics[width=0.9\textwidth]{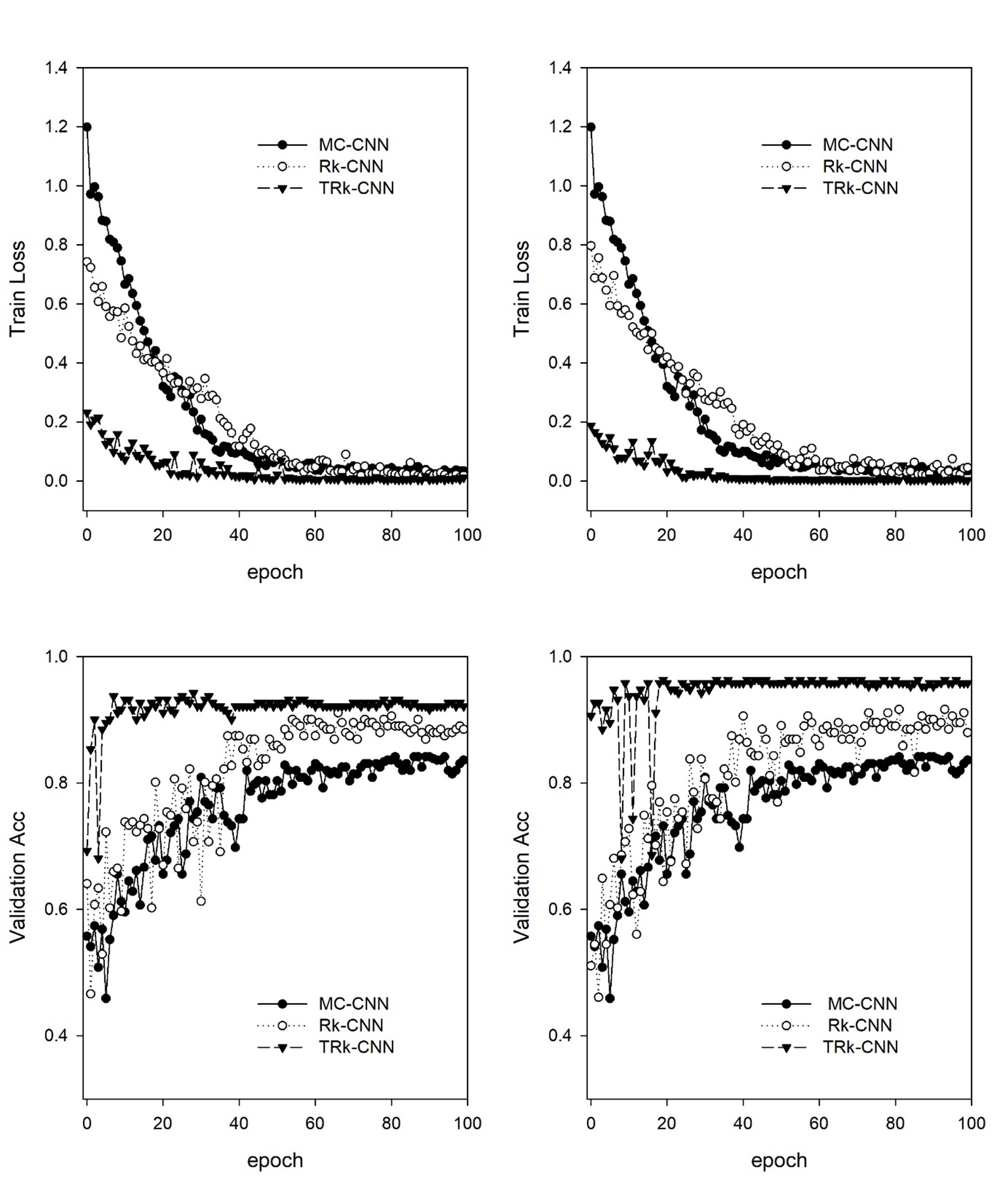}
	\caption{Training loss and validation accuracy of TRk-CNN, RK-CNN, and MC-CNN}
	\label{img:trainloss0}
\end{figure}

Figures \ref{img:trainloss0} show the training loss and validation accuracy of the sub-CNN models of Rk-CNN and TRk-CNN during the 100 epochs, along with those of MC-CNN. Since the number of classes to classify is different, there is a limit to directly comparing MC-CNN with sub-CNN models of Rk-CNN and TRk-CNN, in terms of validation accuracy. However, looking at the tendency of the graphs, TRk-CNN's training loss and validation accuracy converge from earlier epochs than the Rk-CNN and MC-CNN. In other words, by exchanging the ROI extracted from different models, additional information on the input is obtained, so that training with lower loss becomes possible. This explains why TRk-CNN performs better than Rk-CNN considering that the total error of Rk-CNN is bound to the max error of sub-model. As a result, the validation accuracy of the sub-CNN model in TRk-CNN is higher than that of Rk-CNN.

\subsection{Evaluation results of optimized TRk-CNN for glaucoma detection}
Table \ref{tab:result_trk} shows the results of optimized TRk-CNN models trained in several different conditions. DISC, EDISC, and ORIGINAL represent three different regions of the fundus image, all of which were trained using \textit{CEALoss}. ENSEMBLE is the result of majority voting on the predicted classes of three models, and if there is no dominant class it follows the result of DISC model. DISC\textsuperscript{1} and DISC\textsuperscript{2} are models for comparison, and DISC\textsuperscript{1} shows the result of training using \textit{CELoss} instead of \textit{CEALoss} in disc region. DISC\textsuperscript{2} is the result of training glaucoma suspect's ROI with \textit{Cam}\textsubscript{1}\textsuperscript{1} +  \textit{Cam}\textsubscript{0}\textsuperscript{2} instead of \textit{Cam}\textsubscript{0}\textsuperscript{1} +  \textit{Cam}\textsubscript{1}\textsuperscript{2}. Since the loss of DISC\textsuperscript{2} uses \textit{CEALoss}, only the difference of performance according to ROI is compared.
\begin{table}[!tbh]
  \begin{center}
    \scalebox{0.8}{
    \begin{tabular}{|l|c|c|c|c|c|c|c|c|}
     \hline
      Method & \textit{Acc}(\%) & \textit{Sp}(\%) & \textit{Se}\textsuperscript{S}(\%) & \textit{Se}\textsuperscript{G}(\%) & \textit{Pr}\textsuperscript{S}(\%) & \textit{Pr}\textsuperscript{G}(\%) & \textit{F1}\textsuperscript{S}(\%) & \textit{F1}\textsuperscript{G}(\%)\\
     \hline\hline
     ENSEMBLE & \textbf{92.96} & \textbf{93.33} & \textbf{95.12} & 91.57 & \textbf{81.25} & \textbf{98.70} & \textbf{87.64} & 95.00\\
     DISC & 91.46 & 89.33 & 90.24 & \textbf{93.98} & 78.72 & 96.30 & 84.09 & \textbf{95.12}\\
     EDISC & 88.94 & 90.67 & 90.24 & 86.75 & 68.52 & 98.63 & 77.89 & 92.31\\
     ORIGINAL & 90.45 & 92.00 & 92.68 & 87,95 & 77.55 & 97.33 & 84.44 & 92.41\\
     \hline
     DISC\textsuperscript{1} & 88.94 & 89.33 & 85.37 & 90.36 & 74.47 & 94.94 & 79.55 & 92.59\\
     DISC\textsuperscript{2} & 86.43 & 86.67 & 78.05 & 90.36 & 86.49 & 86.21 & 82.05 & 88.24\\
     \hline
     \end{tabular}}
  \end{center}
  \caption{Evaluation results of TRk-CNN models for glaucoma detection}
  \label{tab:result_trk}
\end{table}

Since DISC have the highest performance among the models studied in the three regions, we used the disc region in the comparison of TRk-CNN, Rk-CNN, and MC-CNN. The best overall performance was the ENSEMBLE model which achieved the highest results for all metrics except sensitivity and F1-score for glaucoma. In terms of accuracy, ENSEMBLE achieved 92.96\%, which is 1.50\% higher than DISC, 4.02\% higher than EDISC, and 2.51\% higher than ORIGINAL. From the specificity perspective, ENSEMBLE was the highest at 93.33\%, which is 4\% higher than DISC, 2.66\% higher than EDISC, and 1.33\% higher than MC-CNN. The sensitivities of glaucoma suspect for ENSEMBLE was 95.12\%, which is 4.88\% higher than both DISC and EDISC, and 2.44\% higher than ORIGINAL. The precision of glaucoma suspect for ENSEMBLE achieved 81.25\%, which is 2.53\% higher than DISC, 12.73\% higher than EDISC, and 3.70\% higher than ORIGINAL. Considering the trade-off between sensitivity and precision, F1-score for glaucoma suspect in ENSEMBLE was the highest at 87.64\% which is 3.55\% higher than DISC, 9.75\% higher than EDISC, and 3.20\% higher than ORIGINAL. The sensitivity of glaucoma for ENSEMBLE was 91.57\% while DISC was 93.98\%, EDISC was 86.75\%, and ORIGINAL was 87.95\%. The precision of glaucoma for TRk-CNN achieved 98.70\% while 96.30\% for DISC, 98.63\% for EDISC, and 97.33\% for ORIGINAL. Finally, F1-score of glaucoma for DISC was the highest at 95.12\% which is 0.12\% higher than ENSEMBLE, 2.81\% higher than EDISC, and 2.71\% higher than ORIGINAL. The results show that referring to the disc region is the best performance for specifying glaucoma. However, in the case of detecting normal and glaucoma suspect eyes, it is better to refer to a wider area, and as a result, the ENSEMBLE model that combines all of these is the best. 

From the results of DISC and DISC\textsuperscript{1}, using \textit{CEALoss} instead of \textit{CELoss} showed higher performance in all metrics. This means that lower \textit{CELoss} does not necessarily result in higher accuracy as we explained earlier. Also, if we use a metric other than accuracy as an evaluation, many variations are possible. For example, since the Dice Similarity Coefficient score (\textit{DCS}) is the main metric for the segmentation problem, the combined loss of \textit{CELoss} and \textit{DCS} may show better performance.

The results of DISC and DISC\textsuperscript{2} showed that the performance of DISC was higher in all the indicators except precision for glaucoma suspect. However, the F1-score for glaucoma suspect was higher on the DISC, so overall it was better to use ROI as \textit{Cam}\textsubscript{0}\textsuperscript{1} +  \textit{Cam}\textsubscript{1}\textsuperscript{2}. As described in earlier section, defining ROI as \textit{Cam}\textsubscript{0}\textsuperscript{1} +  \textit{Cam}\textsubscript{1}\textsuperscript{2} is considered to contain information that is likely to be the opposite of the prediction in each sub-model. In other words, to output the glaucoma suspect class in the primitive classification step of TRk-CNN, the probability of predicting 1 in \textit{Sub}\textsuperscript{1} and 0 in \textit{Sub}\textsuperscript{2} is higher than in the opposite case. Therefore, it is expected that \textit{Cam}\textsubscript{0}\textsuperscript{1} and \textit{Cam}\textsubscript{1}\textsuperscript{2} are highly contrary to predicted class information, and combining these two can transfer more features to the final classification step.

Table \ref{table:final_result} compares the results of previous studies with the results of our proposed TRk-CNN model for glaucoma detection. However, since previous studies were binary classifications that classify normal and glaucoma instead of three classes, we included the binary classification results of the proposed model. Proposed\textsuperscript{1} classifies normal eye as 0, glaucoma suspicious and glaucoma eyes as 1. In other words, Proposed\textsuperscript{1} is the ensemble of \textit{Sub}\textsuperscript{1} models from DISC, EDISC, and ORIGINAL. 
\begin{table}[!tbh]
  \begin{center}
    \scalebox{0.8}{
    \begin{tabular}{|l|c|p{11mm}|p{10mm}|c|c|c|c|c|}
     \hline
      Method & \textit{Acc}(\%) & \textit{Sp}(\%) & \textit{Se}\textsuperscript{S}(\%) & \textit{Se}\textsuperscript{G}(\%) & \textit{Pr}\textsuperscript{S}(\%) & \textit{Pr}\textsuperscript{G}(\%) & \textit{F1}\textsuperscript{S}(\%) & \textit{F1}\textsuperscript{G}(\%)\\
     \hline\hline
     Proposed & \textbf{92.96} & \textbf{93.33} & \textbf{95.12} & 91.57 & 81.25 & \textbf{98.70} & \textbf{87.64} & 95.00\\
     Rk-CNN & 84.92 & 85.33 & 85.37 & 84.34 & 60.34 & \textbf{100.0} & 70.71 & 91.50\\
     MC-CNN & 83.42 & 85.33 & 68.29 & 89.16 & 75.68 & 85.06 & 71.79 & 87.06\\
     \hline\hline
      & Year & Data & \textit{AUC} & \textit{Acc}(\%) & \textit{Sp}(\%) & \textit{Se}(\%) & \textit{Pr}(\%) & \textit{F1}(\%)\\
     \hline\hline
     Proposed\textsuperscript{1} & 2019 & 992 & 0.974 & 95.48 & 93.33 & 96.77 & 96.00 & 96.39\\
     Li \cite{ch2_li2} & 2018 & 48116 & 0.986 & - & 92 & 95.6 & - & -\\
     Fu \cite{ch2_fu} & 2018 & SCES\newline SINDI & 0.918\newline 0.817 & - & - & - & - & -\\
     Li \cite{ch2_li1} & 2016 & ORIGA & 0.838 & - & - & - & - & -\\
     Chen \cite{ch2_chen} & 2015 & ORIGA\newline SCES & 0.831\newline 0.887 & - & - & - & - & -\\
     Dua \cite{ch2_dua} & 2012 & 60 & - & 93.33 & - & - & - & -\\
     Acharya \cite{ch2_acharya} & 2011 & 60 & - & 91.7 & - & - & - & -\\
     Bock \cite{ch2_bock} & 2010 & 575 & 0.88 & - & 85 & 73 & - & -\\
     Nayak \cite{ch2_nayak} & 2009 & 61 & - & - & 80 & 100 & - & -\\
     \hline
     \end{tabular}}
  \end{center}
  \caption{Result table including comparison with results of previous studies}
  \label{table:final_result}
\end{table}

Since TRk-CNN is a model for considering the inter-class relationship, it can be seen that there is no significant difference from using MC-CNN in case of binary classification. This can be seen from the fact that the work of Li \cite{ch2_li2} and the performance of Proposed\textsuperscript{1} do not differ greatly. However, when performing three-class classification, the performance difference between MC-CNN and Proposed is large, because the classes of normal, glaucoma suspect, and glaucoma have a high relation with each other. Therefore, when multi-class classification is performed considering the inter-class relationship, using TRk-CNN can be expected to perform better than the multi-class classification approach.

\begin{figure}[!tbh]
	\centering
	\includegraphics[width=0.85\textwidth]{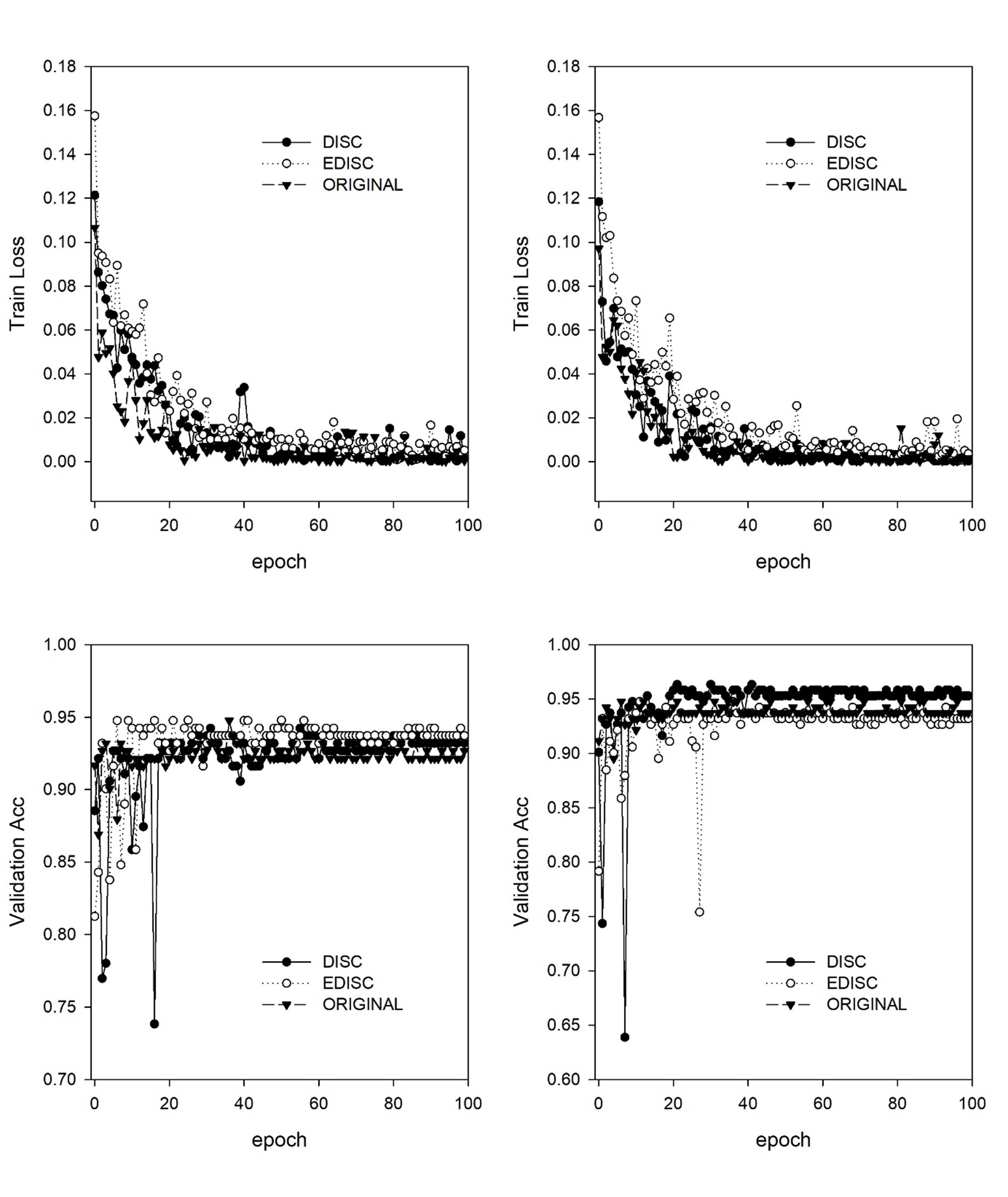}
	\caption{Training loss and validation accuracy of DISC, EDISC, and ORIGINAL}
	\label{img:trainlossopt0}
\end{figure}

Figures \ref{img:trainlossopt0} show the training loss and validation accuracy of 1st and 2nd sub-CNN models of DISC, EDISC, and ORIGINAL, respectively. One notable difference is that the overall validation accuracy of the 1st sub-CNN model was the highest in EDISC, but the results in test-set were the highest in DISC. This implies that the best performance in one sub-model may not necessarily be the best for the aggregated result.

\begin{figure}[!tbh]
	\centering
	\includegraphics[width=0.87\textwidth]{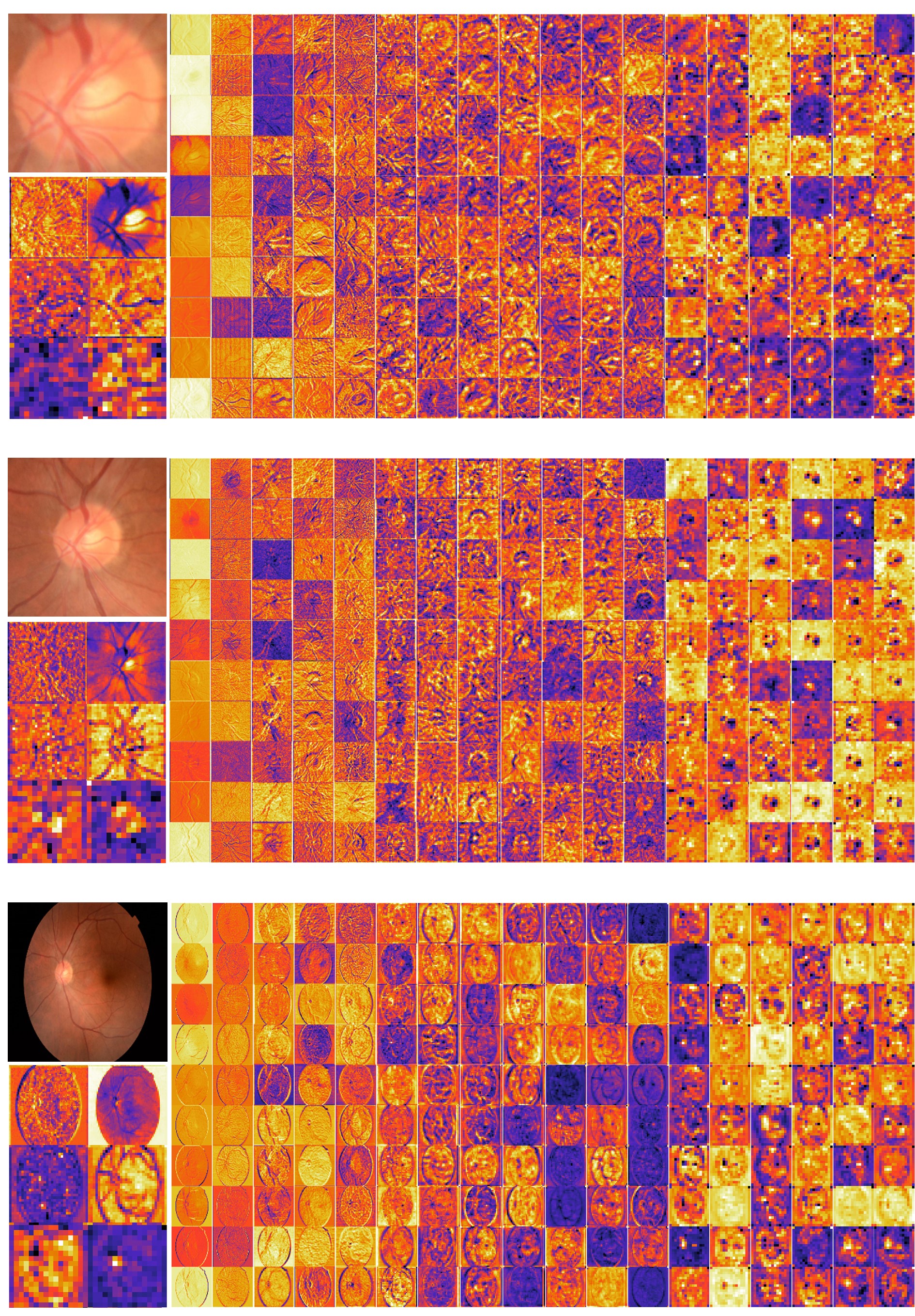}
	\caption{Visualization of the convolution layer in the DISC}
	\label{img:visualize_disc}
\end{figure}

Figures \ref{img:visualize_disc} show how the activation of each convolutional layer is visualized where the input image is the three regions of the same fundus image. The top left image in each figure represents disc, e-disc, and original region for the same fundus image. 

The six images on the bottom left of each figure are visualizations of the activation in the pooling layer of each model. The six images show the deepening of the model from top to bottom, highlighting the retinal blood vessel and disc/cup regions. This can be seen more clearly in DISC and EDISC. Although we manually draw the region box contains only the disc region, we can observe that the model automatically emphasizes the cup region. Other small patch images are from the left to the right in the direction of deepening the model, all visualizing the activation of the convolutional layer. The patches in the same column represent the first 10 filters of the convolutional layer. As shown in the figures, the early part of the convolutional layer extracts low-level feature such as image outline and contrast. As the model deepens, we can see that high-level features are extracted. One peculiar point is that in the case of EDISC and ORIGINAL, the disc region is still emphasized even though the depth of the model is deep enough.

%-------------------------------------------------------------------------
% Major Contribution과 evaluation 결과, future work등을 pharaprasing하면 된다 
\section{Conclusion}
\label{ch2:conclusion}
Our proposed TRk-CNN is a method that can be effectively applied when the classes of images to be classified show a high correlation with each other. The multi-class classification method based on the softmax function, which is generally used, is not effective in this case because the inter-class relationship is ignored. Although there is a Ranking-CNN that takes into account the ordinal classes, it cannot reflect the inter-class relationship to the final prediction. TRk-CNN, on the other hand, combines the weights of the primitive classification model to reflect the inter-class information to the final classification phase. Through extensive experiments, we show that TRk-CNN is superior to both the multi-class classification method and Ranking-CNN method.

We evaluated TRk-CNN in glaucoma image dataset that was collected and labeled from Korea University Medical Center. Glaucoma dataset was labeled into three classes: normal, glaucoma suspect, and glaucoma eyes. Based on the literature we surveyed, this study is the first to classify three status of glaucoma fundus image dataset into three different classes. We compared the evaluation results of TRk-CNN with multi-class CNN (MC-CNN) and Ranking-CNN (Rk-CNN) using the DenseNet as the backbone CNN model. As a result, TRk-CNN achieved an average accuracy of 92.96\%, specificity of 93.33\%, sensitivity for glaucoma suspect of 95.12\% and sensitivity for glaucoma of 93.98\%. Based on average accuracy, TRk-CNN is 8.04\% and 9.54\% higher than Rk-CNN and MC-CNN and surprisingly 26.83\% higher for sensitivity for suspicious than multi-class CNN.

Our TRk-CNN is expected to be effectively applied to the medical image classification problem where the disease state is continuous and increases in the positive class direction. Therefore, we will apply TRk-CNN to medical images with the above characteristics in future work.

\section*{Acknowledgments}
This research was supported by International Research \& Development Program of the National Research Foundation of Korea(NRF) funded by the Ministry of Science, ICT\&Future Planning of Korea(2016K1A3A7A03952054) and by a Korea University Grant (K1625491, K1722121, and K1811051) and by Basic Science Research Program through the National Research Foundation of Korea (NRF) funded by the Ministry of Science and ICT (2018R1C1B6002794). The second funding source had no role in the design or conduct of this research.

\section*{References}

\bibliography{mybibfile}

\end{document}